\documentclass[11pt]{article}

\usepackage[preprint]{acl}

\usepackage{colortbl}
\usepackage{xcolor}

\usepackage{times}
\usepackage{latexsym}

\usepackage[T1]{fontenc}

\usepackage[utf8]{inputenc}

\usepackage{microtype}

\usepackage{inconsolata}

\usepackage{graphicx}

\usepackage{ulem} 
\usepackage{array}
\usepackage{booktabs,amsfonts,dcolumn}
\usepackage{multirow}
\usepackage{arydshln}
\usepackage{tabularx}
\usepackage{amsmath}

\usepackage{algorithm}
\usepackage{algorithmic}

\usepackage{enumitem}
\usepackage{tabularx}

\newcommand{\INPUT}{\item[\textbf{Input:}]}
\newcommand{\OUTPUT}{\item[\textbf{Output:}]}

\title{ExpSeek: Self-Triggered Experience Seeking for Web Agents}

\author{
    Wenyuan Zhang$^{1,2}$, Xinghua Zhang$^{3}$, Haiyang Yu$^{3}$, Shuaiyi Nie$^{1,2}$, \\ \textbf{Bingli Wu}$^{3}$, \textbf{Juwei Yue}$^{1,2}$, \textbf{Tingwen Liu}$^{1,2*}$, \textbf{Yongbin Li}$^{3}$\thanks{corresponding authors.} \\
  $^1$Institute of Information Engineering, Chinese Academy of Sciences \\
  $^2$School of Cyber Security, University of Chinese Academy of Sciences \\
  $^3$Tongyi Lab\includegraphics[height=12pt]{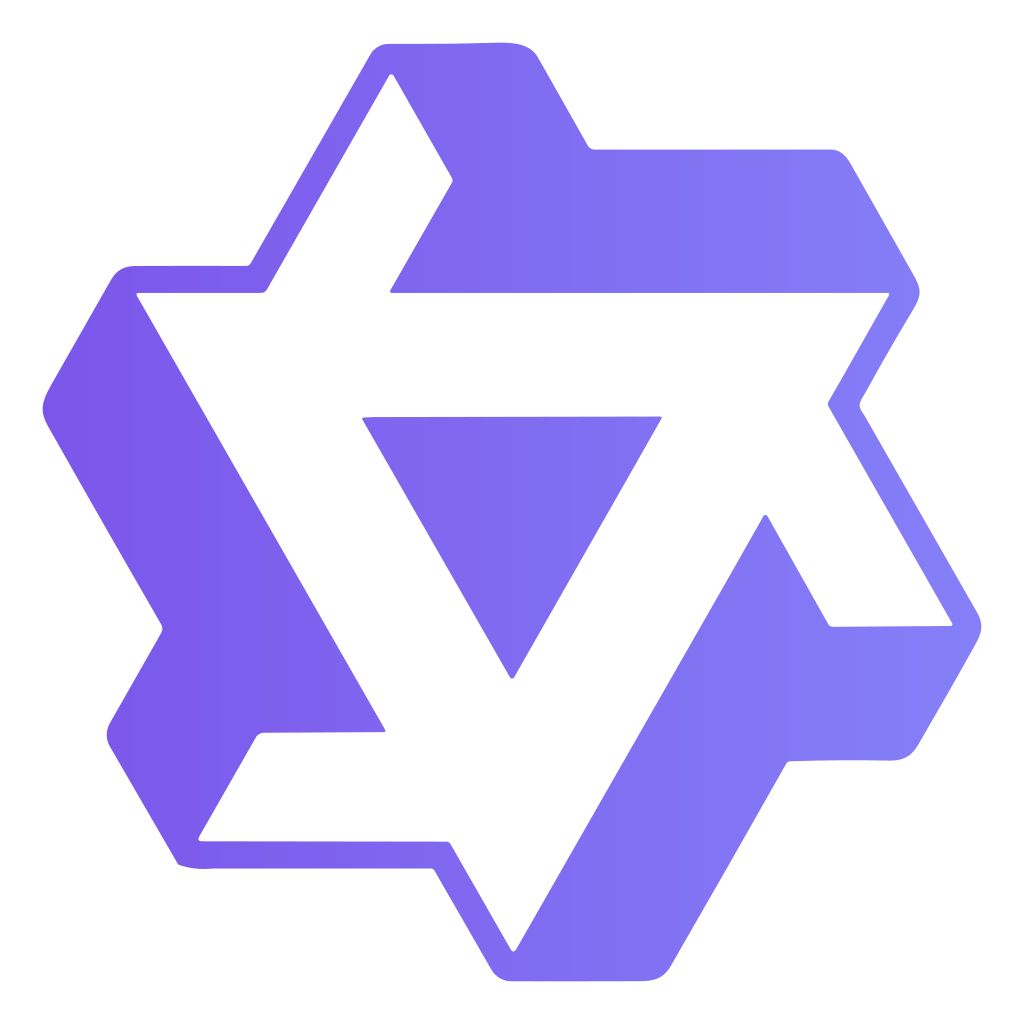}, Alibaba Group \\
  \small\texttt{\{zhangwenyuan, liutingwen\}@iie.ac.cn}, ~
  \small\texttt{\{zhangxinghua.zxh, shuide.lyb\}@alibaba-inc.com}\\
}

\begin{document}
\maketitle

\begin{abstract}
Experience intervention in web agents emerges as a promising technical paradigm, enhancing agent interaction capabilities by providing valuable insights from accumulated experiences. However, existing methods predominantly inject experience passively as global context before task execution, struggling to adapt to dynamically changing contextual observations during agent-environment interaction. We propose \textbf{ExpSeek}, which shifts experience toward step-level proactive seeking: (1) estimating step-level entropy thresholds to determine intervention timing using the model's intrinsic signals; (2) designing step-level tailored experience content. Experiments on Qwen3-8B and 32B models across four challenging web agent benchmarks demonstrate that ExpSeek achieves absolute improvements of 9.3\% and 7.5\%, respectively. Our experiments validate the feasibility and advantages of entropy as a self-triggering signal, reveal that even a small-scale 4B experience model can significantly boost the performance of larger agent models. The code is released at \url{https://github.com/WYRipple/ExpSeek}.
\end{abstract}

\section{Introduction}



Advances in large language models (LLMs) are gradually unlocking greater potential for agents~\cite{team2025kimi,qu2025tool,zhang2025survey}.
Recently, web agents powered by search engines have gained considerable attention for their capability to retrieve relevant information from the web and address complex user queries~\cite{WebAgents_survey,song-etal-2025-beyond}.
Accordingly, the agent needs to possess the ability to conduct multi-turn interactions with the web to obtain evidence~\cite{webagentR1}, and leverage it to provide answers to the user's queries~\cite{lu2024weblinx}.
However, the open web is noisy and partially observable with sparse useful evidence, posing challenges to the agent's reliability~\cite{leelearning}.
Agents powered by LLMs, particularly small-scale, cost‑effective models, often exhibit inefficient exploration in multi‑turn interactions with the environment or respond prematurely, resulting in unreliable answers~\cite{gao2025beyond}.

\begin{figure}[!t]  
\centering  
\includegraphics[width=7.7cm]{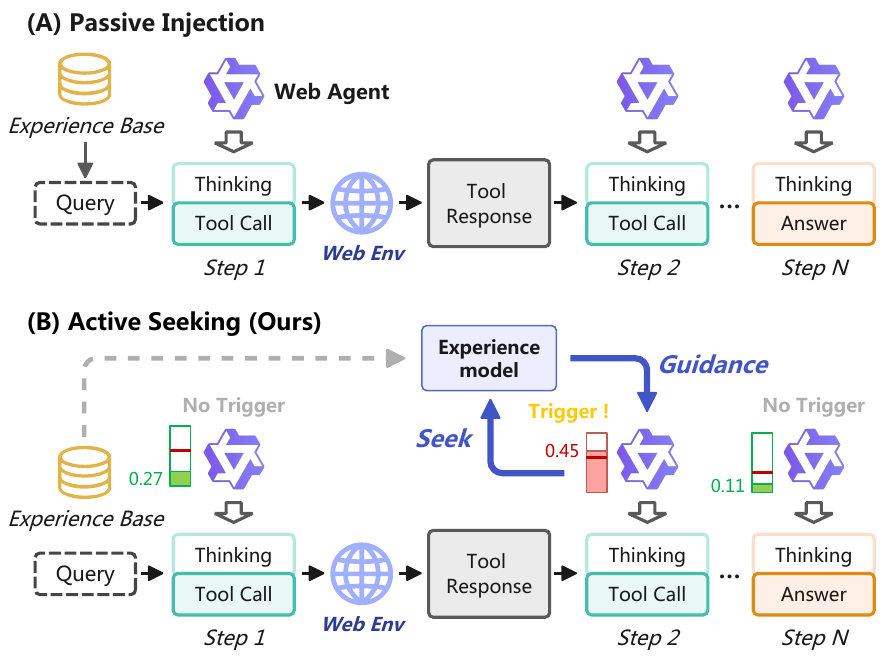}  
\caption{Comparison of experience intervention frameworks. Panel A shows the traditional global passive injection of experience, while we extend the framework to Panel B, where the agent proactively seeks guidance at each step based on its own signals.}
\label{fig:motivation}
\end{figure}

As demonstrated by previous studies~\cite{ExpeL,ICLR-Synapse}, agents, like humans~\cite{Google-welcome}, can learn efficiently from experience, drawing on information accumulated from historical interaction trajectories.
Existing methods mainly follow two lines: offline refinement, which post-processes trajectories into reusable patterns and retrieves them at inference time~\cite{ACL24-RaDA,ACL-ExpeTrans}; and online self-evolution, which accumulates experience through iterative interaction and feedback~\cite{ICML-AWM, ACL-contextual-yaoshunyu,EMNLP-evolvesearch}.
Despite their effectiveness, the experience is often passively injected as a global context before the task execution, as shown in Figure~\ref{fig:motivation} (A). 
However, during the agent’s interaction with the environment, the contextual observations continuously change. If the agent relies solely on the initial context without proactively acquiring and integrating fresh experience, its decision-making may become suboptimal or even misaligned with the current situation.
Compared with passively injecting experience, {\it why not empower the agent to proactively seek experience during its interaction with the environment for more precise guidance}?

This paper proposes \textbf{ExpSeek}, a self-triggered experience-seeking framework that clarifies {\it when to seek experiences} and {\it which ones to seek}, as briefly depicted in Figure~\ref{fig:motivation} (B):

(1) {\it \textbf{The timing for seeking experience}}:
Frequent or sparse experience seeking makes it difficult to achieve ideal results in terms of effectiveness and efficiency.
An optimal timing for seeking experience is when the agent becomes confused and genuinely in need of guidance during interaction.
To this end, we estimate a promising threshold interval for each step based on entropy value through logistic regression and bootstrap resampling, and use the estimated interval to control the timing of experience seeking.
(2) {\it \textbf{The concrete content of experience}}:
After determining the timing for seeking experience, the next step is to decide its concrete content.
Firstly, we formalize an experience base from pairs of successful and failed trajectories.
Each triplet contains the erroneous behavior, mistake analysis, and corrective cues, and is grouped by topic.
Subsequently, the experience model reads the historical context of the current step, retrieves the related experience triplets from the experience base, and generates guidance tailored to the ongoing interaction.

We evaluate ExpSeek on four challenging web agent benchmarks using agents powered by Qwen3-8B and Qwen3-32B. ExpSeek respectively achieves 9.3\% and 7.5\% absolute improvements over the 8B and 32B base models, and outperforms passive experience injection baselines by 6.7\% and 6.0\%, validating the effectiveness of proactive experience seeking during interaction with environment. 
Extensive analyses reveal some key insights: (1)
The entropy of a model itself is an effective trigger signal and can indicate whether it should receive experience guidance, controlling the timing for seeking experience.
(2) Even with a small‑scale 4B model, the proactive experience seeking paradigm of ExpSeek can still yield significant performance gains for a slightly larger 32B model.
(3) The self-triggered experience seeking increases the entropy of the model during the intermediate steps of interaction with the environment, while decreasing entropy at the final answer step, confirming the enhanced exploration and ultimately enabling more effective convergence toward the correct answer.

In summary, our contributions are as follows:
\begin{itemize}[nosep, topsep=2pt]

    \item We propose \textbf{ExpSeek}, a self-triggered experience seeking framework, inspiring a proactive paradigm for seeking experience distinct from passive experience injection.

    \item We explore and confirm that the entropy of a model itself can serve as an intrinsic signal of the timing for proactively seeking experience. Additionally, we build the experience base with experience triplets, and an experience model is designed to dynamically generate experience guidance during agent–environment interactions, based on experience triplets and historical context.
    
    \item Extensive experiments and analyses show the significant advantages of ExpSeek with average improvements of 9.3\% and 7.5\% on 8B/32B models, an increase of up to 14.6\%.
    
\end{itemize}

\section{Related Work}
In this section, we discuss the most relevant lines of work. A broader discussion of related work is provided in Appendix~\ref{app:extended_rw}.

\subsection{Experience Intervenes in Agents}

Experience~\cite{Google-welcome} serves as long-term memory to prevent repeated mistakes and accumulate insights, distinct from short-term contextual memory~\cite{GuiBin-memory}.

Recent work can be categorized into two lines:
(1) \textbf{Offline Experience} refines experience from offline training sets. Early methods directly use raw trajectories~\cite{ICLR-Synapse}, while later work structures experience~\cite{ExpeL, ACL24-RaDA, tongyi-memp, brew} and induces patterns from successful and failed trajectories~\cite{TF-GRPO} to extract transferable~\cite{ACL-ExpeTrans, agentkb} and reusable reasoning units.
(2) \textbf{Self-Evolution} accumulates domain-specific experience online through gradient-free training~\cite{seed-flex} or by shifting the model's output distribution~\cite{ACL-Browsing-Like-Human}, with real-time updates~\cite{ICML-AWM, ACL-contextual-yaoshunyu, learning-on-the-job, tongyi-remember-me, EMNLP-evolvesearch, reasoningbank, EMNLP-Self-Guided}.
%
However, such passive experience injection is difficult to align with step decisions.

\subsection{Entropy in LLM Reasoning}

Entropy serves as a fundamental uncertainty metric widely used for static reasoning evaluation, including answer selection~\cite{pmlr-Self-Evaluation,TMLR-CoG}, confidence calibration~\cite{ACL-quantifying}, and error detection~\cite{Nature-semantic-entropy,kdd-uncertainty-survey}. 
As reasoning scales to multi-step paradigms where responses are decomposed into atomic steps~\cite{zhang2025s1bench,nie2026attnpo,ai2026shapes}, the role of entropy extends to training. 
Particularly in reinforcement learning, entropy not only reflects sampling diversity to facilitate exploration~\cite{Neurips-8020rule,first-return}, but also acts as a fine-grained signal for step-level credit assignment in long-horizon reasoning~\cite{Harnessing,ACL-offline}, and further extends to incentivize exploration across multi-turn interactions~\cite{ARPO}. Recent work also investigates entropy collapse during RLVR 
training and proposes mechanisms to regulate entropy 
dynamics~\cite{jin2026revisiting}.
Confidence-based triggering has been explored in 
retrieval-augmented settings~\cite{ding2026exdr}.
This demonstrates the potential of entropy as a self-trigger 
signal for web agents.

\section{Preliminaries}

\noindent \textbf{Agent Framework.}
Following \citet{wu-etal-2025-webwalker}, we construct web agents based on the ReAct framework~\cite{yao2023react}, modeling problem-solving as an interleaved sequence of reasoning and acting. A trajectory $\tau$ is defined as:
\begin{equation}
\tau=(q, R_1, O_1, \dots, R_t, O_t, \dots, R_T),
\end{equation}
where $q$ is the query, $R_t$ is the agent's response at step $t$, and $O_t$ is the environment observation. We distinguish two types of steps:

\begin{itemize}[leftmargin=*, nosep, topsep=2pt]
    \item \textbf{Process Step} ($S^p_t = (R_t, O_t)$, $t < T$): The response $R_t = \langle z_t, a_t \rangle$ contains reasoning thoughts $z_t$ (enclosed in \texttt{<think>} tags) and an action $a_t$ (in \texttt{<tool\_call>} tags), followed by the tool's response $O_t$.

    \item \textbf{Answer Step} ($S^a_T = R_T$): The terminal step produces $R_T = \langle z_T, \texttt{y} \rangle$, where the final answer \texttt{y} is wrapped in \texttt{<answer>} tags.
\end{itemize}

\noindent \textbf{Step Entropy.}
To quantify the agent's confidence in each step of interaction, we compute the step entropy as the average token entropy across each response $R_t$. Specifically, the entropy of the $i$-th token $x_i$ with preceding context $h_i$ is defined as $H(x_i)=-\sum_{v\in \mathcal{V}}P(v\mid h_i)\log{P(v\mid h_i)}$, where $\mathcal{V}$ is the vocabulary and $P(\cdot \mid h_i)$ is the model's predicted distribution. Step entropy is computed:
\begin{equation}
    \bar{H}_t = \frac{1}{|R_t|} \sum_{x \in R_t} H(x),
\end{equation}
where $|R_t|$ denotes the number of tokens in $R_t$. 
%

\noindent\textbf{Experience Intervention.}
We formally define experience intervention as a class of methods comprising two phases:
\begin{itemize}[leftmargin=*, nosep, topsep=2pt]
    \item \textbf{Construction Phase}: Given a training corpus $\mathcal{D}_{train}=\{(q_i, \tau_i, y_i)\}_{i=1}^N$ with queries $q_i$, trajectories $\tau_i$, and ground truth $y_i$, experience acquisition is formalized as $\mathcal{E} = \mathcal{F}(\mathcal{D}_{train})$, where $\mathcal{F}$ is a function that extracts an experience base $\mathcal{E}$ from $\mathcal{D}_{train}$\footnote{Self-evolution methods have the same $\mathcal{D}_{train}$ and $\mathcal{D}_{test}$, but typically do not provide ground truth $y$.}.

    \item \textbf{Utilization Phase}: At step $t$ with context $h_t \!=\! (q,\ldots, R_t, O_t)$, the agent obtains applicable experience $e_t \! =\! \mathcal{G}(\mathcal{E}, h_t)$ through mapping function $\mathcal{G}$, which serves as additional input for reasoning.
\end{itemize}
Traditional methods represent a special case where experience is concatenated at the beginning of reasoning, with $e = \mathcal{G}(\mathcal{E}, q)$.
The function $\mathcal{G}$ typically returns the entire base or retrieves relevant cases, providing static experience.
In contrast, our method invokes $\mathcal{G}(\mathcal{E}, h_t)$ at any step $t$ to allow the agent to seek appropriate guidance.

\section{Methodology}

%

This section formally introduces ExpSeek, elaborating on three key components: experience base construction (\S~\ref{sec4-2:exp-cons}),  self-triggering mechanism (\S~\ref{sec4-3:entropy}), and step-level guidance (\S~\ref{sec4-4:framework}).

\begin{figure*}[!t]
\centering  
\includegraphics[width=15cm]{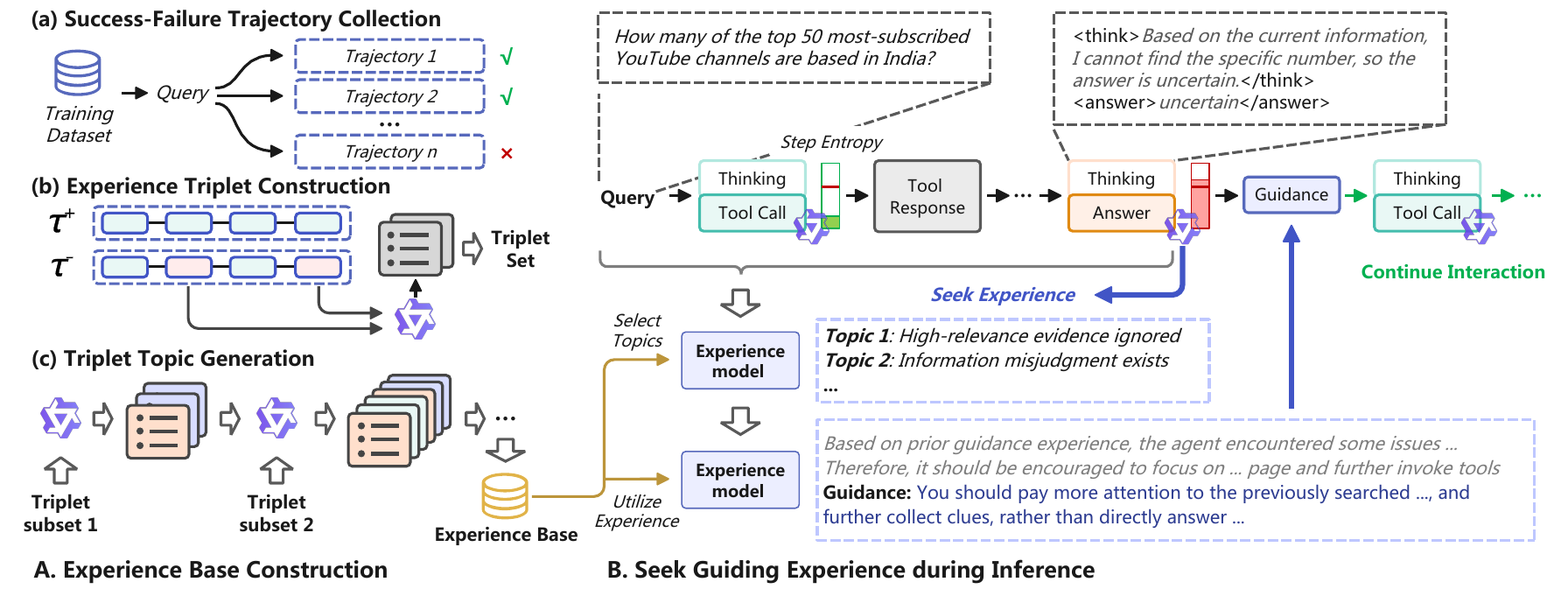}
\caption{The overall architecture of ExpSeek, including experience base construction and actively seeking experience guidance during inference. The step entropy threshold calculation process is not depicted here.}
\label{fig:method}
\end{figure*}

\subsection{Experience Base Construction}
\label{sec4-2:exp-cons}
The core of guiding experience lies in recreating the problem behavior and simulating guidance.

\noindent \textbf{Guiding Experience Schema.} We design experience triplets containing: (1) \textit{Behavior}: objectively describes the state and action at the current step; (2) \textit{Mistake}: identifies errors by contrasting with correct trajectories; (3) \textit{Guidance}: provides directional guidance based on error analysis, without directly offering answers or specific clues.

\noindent \textbf{Construction Process.}
As shown in Figure~\ref{fig:method} (A), construction involves three steps. 
%
First, for each query in $\mathcal{D}_{train}$, we sample $k$ trajectories with the agent model and pair successful and failed ones.
%
%
Second, for each trajectory pair $(\tau^+, \tau^-)$, we use a tool model to analyze the failed trajectory $\tau^- = \{S_1, S_2, \ldots, S_T\}$ by contrasting with the successful one $\tau^+$. The tool model assigns each step $S_t$ a binary correctness label $y_t \in \{0, 1\}$ and outputs triplets for incorrect steps (where $y_t=0$).
Finally, we prompt the tool model to induce topics for triplets using an iterative batch processing approach: when processing each new batch, the model takes all previously generated triplets with their assigned topics as input, then either assigns existing topics, modifies them, or creates new topics for the current batch.
%
This yields a guiding experience base $\mathcal{E}$ organized into topics, with separate collections $\mathcal{E}_p$ and $\mathcal{E}_a$ for process and answer steps respectively.

\subsection{Entropy as Self-Trigger}
\label{sec4-3:entropy}

\subsubsection{Entropy Analysis for Step Correctness}
Determining when to trigger intervention is challenging. Using reward models to analyze each step incurs prohibitive costs. 
Inspired by prior research on entropy-based error detection and confidence calibration~\cite{ACL-quantifying,kdd-uncertainty-survey}, we hypothesize that the step entropy of web agents has the potential to reflect the agent's inherent state.
%
%
We focus on (1) whether entropy can distinguish correct from incorrect steps in web agent reasoning, and (2) whether this distinguishability differs between process and answer steps.

We analyze training trajectories from \S~\ref{sec4-2:exp-cons}. 
Based on correctness labels $y_t$ assigned during trajectory pairing, we construct two step collections:
\begin{equation}
\begin{aligned}
\mathcal{S}^+ &= \{S_t \in \tau^+\} \cup \{S_t \in \tau^- \mid y_t = 1\}, \\
\mathcal{S}^- &= \{S_t \in \tau^- \mid y_t = 0\},
\end{aligned}
\end{equation}
where $\mathcal{S}^+$ aggregates all steps from successful trajectories and correctly-labeled steps from incorrect trajectories, while $\mathcal{S}^-$ comprises only the incorrect steps from failed trajectories. We further partition these collections by step type into process steps ($\mathcal{S}^+_p, \mathcal{S}^-_p$) and answer steps ($\mathcal{S}^+_a, \mathcal{S}^-_a$), then analyze their entropy distributions $\bar{H}_t$.
%

\begin{figure}[!t]  
\centering  
\includegraphics[width=7.7cm]{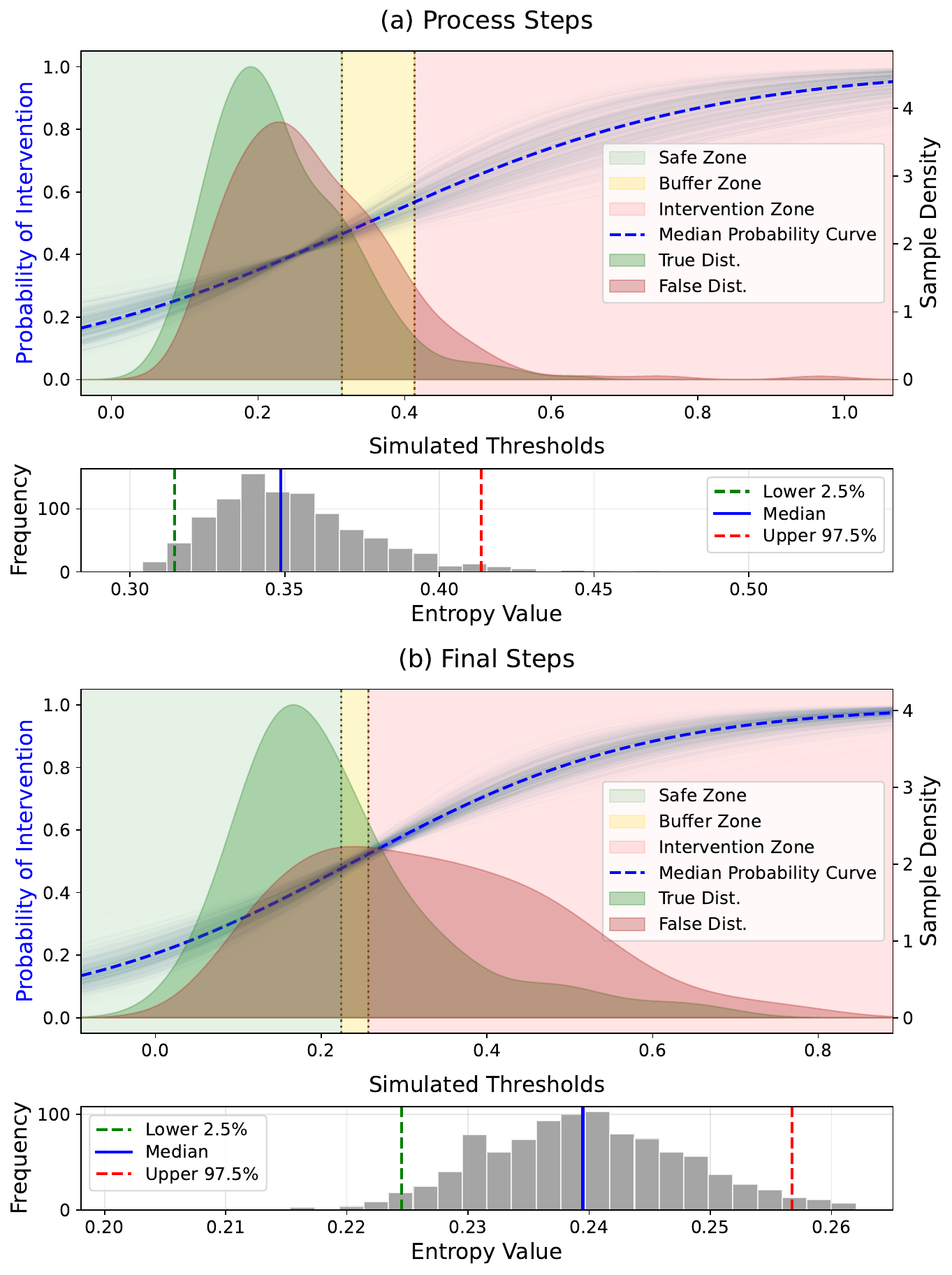}  
\caption{Entropy distributions of process and answer steps on $\mathcal{D}_{train}$ for Qwen3-8B, with fitted logistic regression curves. Green zone indicates no intervention during inference, red indicates intervention, and yellow indicates probabilistic intervention.}
\label{fig:method_entropy_8b}
\end{figure}

Figure~\ref{fig:method_entropy_8b} shows that $\mathcal{S}^+$ consistently exhibits lower entropy than $\mathcal{S}^-$ for both step types.
The Kolmogorov-Smirnov test~\cite{berger2014kolmogorov} confirms that the distributions of correct and incorrect steps are statistically separable for both types (process steps: KS=0.1998, p<0.001; answer steps: KS=0.3809, p<0.001).
However, separability differs substantially in practice: process steps show considerable overlap as agents naturally explore multiple paths, producing high entropy even when correct (AUC=0.6223, indicating weak discrimination~\cite{bradley1997use}), while answer steps demonstrate much clearer separation (AUC=0.7187, indicating acceptable discrimination).
This suggests that entropy serves as a noisy but valid signal for process steps, and becomes more reliable for answer steps, motivating us to estimate thresholds for triggering intervention.

\subsubsection{Threshold-Based Triggering}

\noindent \textbf{Threshold Estimation via Bootstrap.}
To quantify when intervention should occur, we estimate threshold intervals that capture uncertainty in distinguishing correct from incorrect steps. We formulate this as binary classification: given step entropy $\bar{H}_t$, predict correctness $y_t \in \{0, 1\}$.

We fit separate logistic regression models for process steps ($\mathcal{S}^+_p, \mathcal{S}^-_p$) and answer steps ($\mathcal{S}^+_a, \mathcal{S}^-_a$), where each model learns:
\begin{equation}
P(y_t=0 \mid \bar{H}_t) = \frac{1}{1 + e^{-(w \cdot \bar{H}_t + b)}},
\end{equation}
modeling the probability of incorrectness, where higher entropy corresponds to higher error probability. The decision boundary at $P=0.5$ yields threshold $\theta = -b/w$.

Since a single fitted model on limited data may be unstable, we employ bootstrap resampling: we sample with replacement from the respective step collections (preserving original sizes) $N$ times, fit logistic regression on each sample to obtain $\{\theta^{(i)}\}_{i=1}^N$, and compute the 95\% confidence interval as our threshold range:
\begin{equation}
    \begin{aligned}
    \theta_{\text{lower}} &= Q_{0.025}(\{\theta^{(i)}\}), \\
    \theta_{\text{upper}} &= Q_{0.975}(\{\theta^{(i)}\}),
    \end{aligned}
\end{equation}
where $Q_p$ denotes the $p$-th quantile. This yields separate threshold intervals $[\theta_{\text{lower}}^p, \theta_{\text{upper}}^p]$ for process steps and $[\theta_{\text{lower}}^a, \theta_{\text{upper}}^a]$ for answer steps.
More implementation details are provided in Appendix~\ref{app:threshold}.

\noindent\textbf{Probabilistic Intervention at Inference.}
During inference, for each step $S_t$, we compute its entropy $\bar{H}_t$ and determine the intervention probability based on the corresponding threshold interval (using $[\theta_{\text{lower}}^p, \theta_{\text{upper}}^p]$ for process steps or $[\theta_{\text{lower}}^a, \theta_{\text{upper}}^a]$ for answer steps):
\begin{equation}
p_{\text{intervene}} = 
\begin{cases}
0 & \bar{H}_t < \theta_{\text{lower}} \\[4pt]
\frac{\bar{H}_t - \theta_{\text{lower}}}{\theta_{\text{upper}} - \theta_{\text{lower}}} & \theta_{\text{lower}} \leq \bar{H}_t \leq \theta_{\text{upper}} \\[4pt]
1 & \bar{H}_t > \theta_{\text{upper}}
\end{cases}
\end{equation}
We trigger experience guidance (\S~\ref{sec4-4:framework}) with probability $p_{\text{intervene}}$, where low-entropy steps receive no intervention, high-entropy steps (indicating low confidence) always receive guidance, and intermediate cases are handled probabilistically to balance intervention frequency with agent autonomy.

\begin{table}[!t]
\renewcommand{\arraystretch}{0.9}
\small
\centering
    \begin{tabular*}{0.48\textwidth}{@{\extracolsep{\fill}}l cc cc@{}}
    \toprule
    \textbf{Model} & $\theta_{\text{low}}^p$ & $\theta_{\text{high}}^p$ & $\theta_{\text{low}}^a$ & $\theta_{\text{high}}^a$ \\
    \midrule
    Qwen3-8B  & 0.314 & 0.413 & 0.225 & 0.257 \\
    Qwen3-32B & 0.877 & 1.384 & 0.714 & 0.820 \\
    \bottomrule
    \end{tabular*}
    \caption{Threshold intervals obtained through bootstrap resampling for process and answer steps.}
    \label{table:main_thresholds}
\end{table}

\begin{table*}[!ht]
\renewcommand{\arraystretch}{0.9}
\small
    \centering
    \begin{tabular*}{0.97\textwidth}{@{\extracolsep{\fill}}l cccc ccc c@{}}
    \toprule
    \multirow{2}{*}{\raisebox{-0.5\height}{\textbf{Method}}}
    & \multicolumn{4}{c}{\textbf{WebWalkerQA}} 
    & \multirow{2}{*}{\raisebox{-0.5\height}{\textbf{GAIA}}}
    & \multirow{2}{*}{\raisebox{-0.5\height}{\textbf{\textsc{Seal}}}} 
    & \multirow{2}{*}{\raisebox{-0.5\height}{\textbf{xbench}}}
    & \multirow{2}{*}{\raisebox{-0.5\height}{\textbf{Avg.}}} \\
    \cmidrule(lr){2-5}
    & \texttt{Easy} & \texttt{Medium} & \texttt{Hard} &  Avg. 
    & 
    & 
    & 
    & \\
    \midrule
    \multicolumn{9}{c}{\cellcolor{gray!15}\textit{Qwen3-8B}} \\
    
    No Experience & 42.67\textsubscript{\textcolor{gray}{- 0.0}} & 40.19\textsubscript{\textcolor{gray}{- 0.0}} & 32.56\textsubscript{\textcolor{gray}{- 0.0}} & 38.47\textsubscript{\textcolor{gray}{- 0.0}} & 29.13\textsubscript{\textcolor{gray}{- 0.0}} & 23.23\textsubscript{\textcolor{gray}{- 0.0}} & 25.60\textsubscript{\textcolor{gray}{- 0.0}} & 32.23\textsubscript{\textcolor{gray}{- 0.0}} \\
    Training-Free GRPO & 40.66\textsubscript{\textcolor{red}{$\downarrow$2.0}} & 42.86\textsubscript{\textcolor{blue}{$\uparrow$2.7}} & 38.33\textsubscript{\textcolor{blue}{$\uparrow$5.8}} & 40.62\textsubscript{\textcolor{blue}{$\uparrow$2.2}} & 29.32\textsubscript{\textcolor{blue}{$\uparrow$0.2}} & 25.59\textsubscript{\textcolor{blue}{$\uparrow$2.4}} & 26.00\textsubscript{\textcolor{blue}{$\uparrow$0.4}} & 33.79\textsubscript{\textcolor{blue}{$\uparrow$1.6}} \\
    \textsc{ReasoningBank}$^+$ & 44.50\textsubscript{\textcolor{blue}{$\uparrow$1.8}} & 41.52\textsubscript{\textcolor{blue}{$\uparrow$1.3}} & 36.33\textsubscript{\textcolor{blue}{$\uparrow$3.8}} & 40.78\textsubscript{\textcolor{blue}{$\uparrow$2.3}} & 32.04\textsubscript{\textcolor{blue}{$\uparrow$2.9}} & 26.38\textsubscript{\textcolor{blue}{$\uparrow$3.2}} & 28.00\textsubscript{\textcolor{blue}{$\uparrow$2.4}} & 34.80\textsubscript{\textcolor{blue}{$\uparrow$2.6}} \\
    ExpSeek (Ours) & \textbf{51.00}\textsubscript{\textbf{\textcolor{blue}{$\uparrow$8.3}}} & \textbf{49.52}\textsubscript{\textbf{\textcolor{blue}{$\uparrow$9.3}}} & \textbf{44.22}\textsubscript{\textbf{\textcolor{blue}{$\uparrow$11.7}}} & \textbf{48.25}\textsubscript{\textbf{\textcolor{blue}{$\uparrow$9.8}}} & \textbf{36.89}\textsubscript{\textbf{\textcolor{blue}{$\uparrow$7.8}}} & \textbf{30.16}\textsubscript{\textbf{\textcolor{blue}{$\uparrow$6.9}}} & \textbf{37.20}\textsubscript{\textbf{\textcolor{blue}{$\uparrow$11.6}}} & \textbf{41.50}\textsubscript{\textbf{\textcolor{blue}{$\uparrow$9.3}}} \\
    - only process step & 44.17\textsubscript{\textcolor{blue}{$\uparrow$1.5}} & 47.91\textsubscript{\textcolor{blue}{$\uparrow$7.7}} & 38.89\textsubscript{\textcolor{blue}{$\uparrow$6.3}} & 43.66\textsubscript{\textcolor{blue}{$\uparrow$5.2}} & 33.01\textsubscript{\textcolor{blue}{$\uparrow$3.9}} & 27.17\textsubscript{\textcolor{blue}{$\uparrow$3.9}} & 28.40\textsubscript{\textcolor{blue}{$\uparrow$2.8}} & 36.59\textsubscript{\textcolor{blue}{$\uparrow$4.4}} \\
    - only answer step & 49.50\textsubscript{\textcolor{blue}{$\uparrow$6.8}} & 48.48\textsubscript{\textcolor{blue}{$\uparrow$8.3}} & 41.22\textsubscript{\textcolor{blue}{$\uparrow$8.7}} & 46.40\textsubscript{\textcolor{blue}{$\uparrow$7.9}} & 30.29\textsubscript{\textcolor{blue}{$\uparrow$1.2}} & 30.08\textsubscript{\textcolor{blue}{$\uparrow$6.9}} & 34.80\textsubscript{\textcolor{blue}{$\uparrow$9.2}} & 39.06\textsubscript{\textcolor{blue}{$\uparrow$6.8}} \\

    \midrule
    \multicolumn{9}{c}{\cellcolor{gray!15}\textit{Qwen3-32B}} \\
    No Experience & 46.67\textsubscript{\textcolor{gray}{- 0.0}} & 47.24\textsubscript{\textcolor{gray}{- 0.0}} & 41.11\textsubscript{\textcolor{gray}{- 0.0}} & 45.01\textsubscript{\textcolor{gray}{- 0.0}} & 36.50\textsubscript{\textcolor{gray}{- 0.0}} & 27.80\textsubscript{\textcolor{gray}{- 0.0}} & 27.40\textsubscript{\textcolor{gray}{- 0.0}} & 37.79\textsubscript{\textcolor{gray}{- 0.0}} \\
    Training-Free GRPO & 46.92\textsubscript{\textcolor{blue}{$\uparrow$0.3}} & 46.85\textsubscript{\textcolor{red}{$\downarrow$0.4}} & 41.67\textsubscript{\textcolor{blue}{$\uparrow$0.6}} & 45.15\textsubscript{\textcolor{blue}{$\uparrow$0.1}} & 36.89\textsubscript{\textcolor{blue}{$\uparrow$0.4}} & 28.19\textsubscript{\textcolor{blue}{$\uparrow$0.4}} & 28.20\textsubscript{\textcolor{blue}{$\uparrow$0.8}} & 38.12\textsubscript{\textcolor{blue}{$\uparrow$0.3}} \\
    \textsc{ReasoningBank}$^+$ & 47.50\textsubscript{\textcolor{blue}{$\uparrow$0.8}} & 47.62\textsubscript{\textcolor{blue}{$\uparrow$0.4}} & 41.67\textsubscript{\textcolor{blue}{$\uparrow$0.6}} & 45.60\textsubscript{\textcolor{blue}{$\uparrow$0.6}} & 33.01\textsubscript{\textcolor{red}{$\downarrow$3.5}} & 29.84\textsubscript{\textcolor{blue}{$\uparrow$2.0}} & 36.33\textsubscript{\textcolor{blue}{$\uparrow$8.9}} & 39.33\textsubscript{\textcolor{blue}{$\uparrow$1.5}} \\
    ExpSeek (Ours) & \textbf{54.33}\textsubscript{\textbf{\textcolor{blue}{$\uparrow$7.7}}} & \textbf{53.05}\textsubscript{\textbf{\textcolor{blue}{$\uparrow$5.8}}} & \textbf{45.89}\textsubscript{\textbf{\textcolor{blue}{$\uparrow$4.8}}} & \textbf{51.09}\textsubscript{\textbf{\textcolor{blue}{$\uparrow$6.1}}} & \textbf{43.88}\textsubscript{\textbf{\textcolor{blue}{$\uparrow$7.4}}} & \textbf{32.76}\textsubscript{\textbf{\textcolor{blue}{$\uparrow$5.0}}} & \textbf{42.00}\textsubscript{\textbf{\textcolor{blue}{$\uparrow$14.6}}} & \textbf{45.32}\textsubscript{\textbf{\textcolor{blue}{$\uparrow$7.5}}} \\
    - only process step & 49.17\textsubscript{\textcolor{blue}{$\uparrow$2.5}} & 47.24\textsubscript{\textcolor{gray}{$\pm$0.0}} & 45.56\textsubscript{\textcolor{blue}{$\uparrow$4.5}} & 47.32\textsubscript{\textcolor{blue}{$\uparrow$2.3}} & 40.58\textsubscript{\textcolor{blue}{$\uparrow$4.1}} & 31.50\textsubscript{\textcolor{blue}{$\uparrow$3.7}} & 30.80\textsubscript{\textcolor{blue}{$\uparrow$3.4}} & 40.81\textsubscript{\textcolor{blue}{$\uparrow$3.0}} \\
    - only answer step & 50.83\textsubscript{\textcolor{blue}{$\uparrow$4.2}} & 48.57\textsubscript{\textcolor{blue}{$\uparrow$1.3}} & 42.78\textsubscript{\textcolor{blue}{$\uparrow$1.7}} & 47.39\textsubscript{\textcolor{blue}{$\uparrow$2.4}} & 37.11\textsubscript{\textcolor{blue}{$\uparrow$0.6}} & 32.52\textsubscript{\textcolor{blue}{$\uparrow$4.7}} & 35.40\textsubscript{\textcolor{blue}{$\uparrow$8.0}} & 41.20\textsubscript{\textcolor{blue}{$\uparrow$3.4}} \\

    \bottomrule
    \end{tabular*}
    \caption{Main results of ExpSeek on four benchmarks using two backbone agents. We report mean accuracy (\%) over five independent runs and absolute improvements over vanilla ReAct without experience. We also provide full ablation results for guiding only process or answer steps.}
    \label{table:main_exp}
\end{table*}

\subsection{Guided Intervention at Inference}
\label{sec4-4:framework}

At inference, we implement $\mathcal{G}(\mathcal{E}, h_t)$ through a experience model $\mathcal{M}_{e}$ that dynamically generates contextualized interventions. The process is illustrated in Figure~\ref{fig:method} (B).

For each step $S_t$, we first compute $\bar{H}_t$ and sample from $p_{\text{intervene}}$ (\S~\ref{sec4-3:entropy}). If triggered and no intervention occurred at the previous step, $\mathcal{M}_{e}$ selects the three most relevant topics from $\mathcal{E}_p$ (for process steps) or $\mathcal{E}_a$ (for answer steps) based on context $h_t$, then adapts experiences under selected topics to generate guidance $e_t$ for the current situation.

The generated guidance is injected differently by step type. For process steps $S^p_t$, $e_t$ is appended to $O_t$. For answer steps $S^a_T$, we extend it to $\{R_T, O_T\}$ with $e_t$ as $O_T$, enabling the agent to continue at step $T+1$ (either refining the answer or invoking tools for further reasoning). 

To prevent over-intervention, we disable intervention at step $t+1$ after any intervention at step $t$, allowing the agent to incorporate guidance before receiving further intervention.

\section{Experiments}
\subsection{Setup}
\noindent\textbf{Benchmarks and Metrics.} 
We evaluate on four challenging real-world web agent reasoning benchmarks:  \textbf{GAIA}~\cite{mialon2023gaia}, \textbf{WebWalkerQA}~\cite{wu-etal-2025-webwalker}, \textbf{xbench-DeepSearch}~\cite{chen2025xbenchtrackingagentsproductivity}, and \textbf{\textsc{Seal-Hard}}~\cite{pham2025sealqaraisingbarreasoning}. 
We construct our training set by sampling 25\% from each difficulty level (easy, medium, hard) in WebWalkerQA with random seed 0, yielding 170 examples, with the remainder used as the test set.
We employ the LLM-as-a-Judge approach for evaluation and report average accuracy across five independent runs.

\noindent\textbf{Tool Environment.} 
Agents are equipped with two fundamental tools: (1) \texttt{Search}, which queries a search engine to return relevant URLs with snippets; and (2) \texttt{Visit}, which accesses a specific URL to retrieve its content.

\noindent\textbf{Configuration.}
We set the maximum number of ReAct steps to 30, treating episodes exceeding this limit as failures. We use \texttt{Qwen3-8B} and \texttt{Qwen3-32B} as agents with sampling temperature of 1.0 and top-p of 0.95. 
For the tool model and $\mathcal{M}_e$, we employ \texttt{Qwen3-235B-A22B-Instruct-2507} in the main experiments.
During experience construction, trajectories are sampled five times, and bootstrap sampling uses $N=1000$.
The constructed experience repositories contain: for 8B, $|\mathcal{E}_p|=196$ (17 topics) and $|\mathcal{E}_a|=190$ (11 topics); for 32B, $|\mathcal{E}_p|=276$ (18 topics) and $|\mathcal{E}_a|=143$ (23 topics).
The threshold intervals derived from bootstrap resampling are shown in Table~\ref{table:main_thresholds}.

\noindent \textbf{Baseline.}
We select two typical experience-based methods as baselines. 
Training-Free GRPO~\cite{TF-GRPO} uses semantic advantages to iteratively acquire and update high-quality experiences from offline trajectories, globally leveraging the experience repository at test time. 
\textsc{ReasoningBank}~\cite{reasoningbank} is a self-evolving experience acquisition scheme that accumulates experiences from online tasks and retrieves them from a continuously updated experience repository in the system prompt during subsequent reasoning. We implement an enhanced version \textsc{ReasoningBank}$^+$ using 235B instead of a weaker reasoning agent to generate experiences.
All experimental and setup details are provided in Appendix~\ref{app:Setting}, \ref{app:prompt_detail}.

\begin{figure*}[!t]
\centering  
\includegraphics[width=16cm]{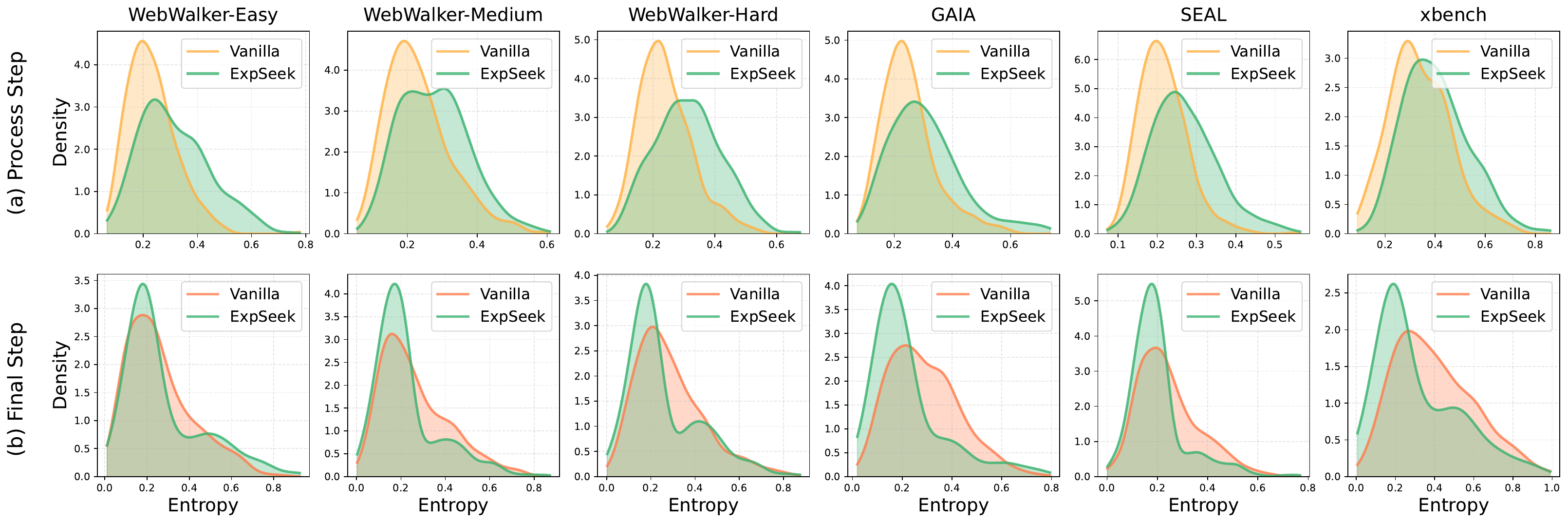}
\caption{Entropy distributions of process and answer steps for Qwen3-8B before and after applying ExpSeek across all benchmarks. Results for Qwen3-32B are provided in Figure~\ref{fig:qwen3_32b_six}.}
\label{fig:qwen3_8b_six}
\end{figure*}

\subsection{Main Results}
Table~\ref{table:main_exp} reports the main experimental results. The key findings are as follows:

(1) \textit{ExpSeek demonstrates significantly superior performance}. ExpSeek achieves average absolute improvements of 9.3\% and 7.5\% over vanilla ReAct on Qwen3-8B and 32B respectively, substantially outperforming baselines across all benchmarks and highlighting the advantages of step-level experience guidance.

(2) \textit{Global intervention methods show limited effectiveness}. Both baseline types struggle with challenging open web reasoning tasks, showing improvements under 3\% or even performance degradation. This indicates that carefully designed global experience injection struggles to adapt to noisy web environments while adding reasoning burden to smaller agents.

(3) \textit{Strong cross-task generalization capability}. Despite being derived entirely from the WebWalkerQA training set, ExpSeek maintains robust performance across three out-of-distribution benchmarks.


We also report pass@3 performance, demonstrating ExpSeek's competitive sampling diversity. Additionally, we provide extensive case studies to intuitively illustrate the effectiveness of our method. Supplementary materials and further discussions can be found in Appendix~\ref{app:Experiment}.

\begin{table}[!t]
\renewcommand{\arraystretch}{0.9}
\small
\centering
    \begin{tabular}{l ccc l}
    \toprule
    \textbf{Method} & \textbf{Acc.} $\uparrow$  & \textbf{Rej.} $\uparrow$ & \textbf{Step} $\downarrow$  & \textbf{Time} $\downarrow$   \\
    \midrule
    \multicolumn{5}{c}{\cellcolor{gray!15}\textit{Qwen3-8B GAIA}} \\
    No Experience & 29.13 & - & 4.47 & 66.94s \\
    Rule + $\mathcal{M}_e$ & 38.81 & 0.0 & 9.52 & 329.71s \\
    Claude-4 + $\mathcal{M}_e$ & 39.47 & 7.4 & 8.55 & 370.82s \\
    Entropy + $\mathcal{M}_e$ & 36.89 & 67.8 & 5.75 & 127.57s \\
    Entropy + \texttt{emb} & 30.92 & 74.2 & 5.54 & 110.61s  \\

    \midrule
    \multicolumn{5}{c}{\cellcolor{gray!15}\textit{Qwen3-8B xbench}} \\
    No Experience & 25.60 & - & 3.31 & 51.06s \\
    Rule + $\mathcal{M}_e$ & 36.40 & 0.0 & 8.52 & 300.09s \\
    Claude-4 + $\mathcal{M}_e$ & 36.60 & 4.8  & 7.40  & 315.39s  \\
    Entropy + $\mathcal{M}_e$ & 37.20 & 42.2  & 5.73  & 143.81s \\
    Entropy + \texttt{emb}  & 27.80 & 44.4 & 4.69 & 88.17s \\
    
    \bottomrule
    \end{tabular}
    \caption{Performance and efficiency with different triggers and retrieval-based guidance in GAIA and xbench. Rej: the proportion of trigger checks that result in non-intervention; Step \& Time: average per question.}
    \label{table:ana_claude4_vs_entropy}
\end{table}

\subsection{Ablation Study}

To validate the individual effectiveness of guiding process and answer steps, we report complete ablation results in Table~\ref{table:main_exp}. Guiding only process or answer steps fails to surpass full intervention, showing significant performance drops on both 8B (-4.91\%, -2.44\%) and 32B (-4.51\%, -4.12\%). 
Notably, guiding only answer steps achieves performance closer to the full method, which validates our observation in \S~\ref{sec4-3:entropy} that answer steps exhibit stronger distributional distinguishability.

\section{Experimental Analyses}

In this section, we conduct an in-depth analysis of ExpSeek across four dimensions: internal mechanisms, design rationale, scalability, and efficiency.

\begin{figure}[t]  
\centering  
\includegraphics[width=7.5cm]{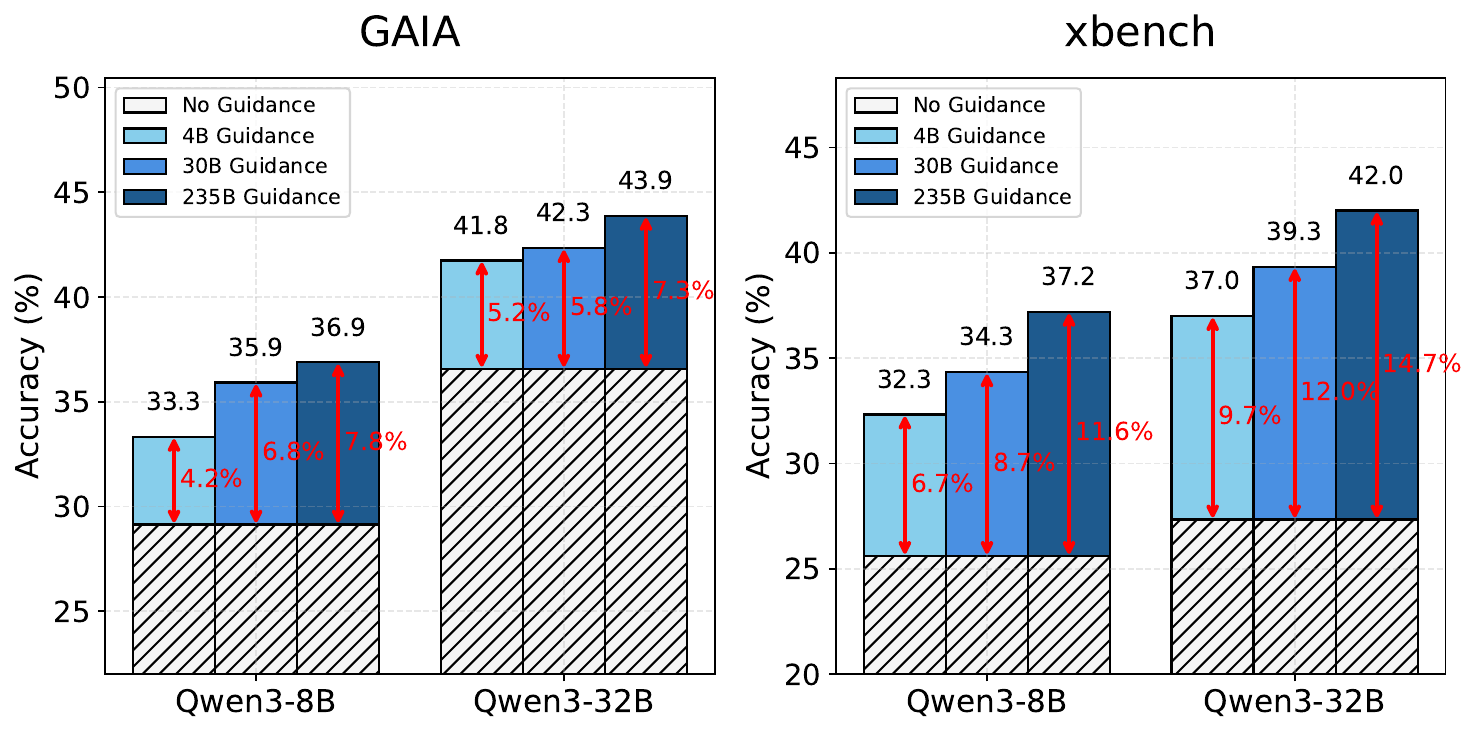}  
\caption{Scaling Law of experience model $\mathcal{M}_e$.}
\label{fig:ana_small}
\end{figure}

\subsection{How Does ExpSeek Work Internally?}
To reveal how ExpSeek works internally, we visualize the entropy distribution shifts of agent outputs before and after experience guidance (Figure~\ref{fig:qwen3_8b_six}). The results show that guidance increases entropy in process steps, enabling the agent to escape local decisions and explore broader reasoning spaces; conversely, the entropy distribution of answer steps shifts toward lower values with a sharper peak, indicating the agent converges to correct answers with higher confidence. This diverge-then-converge behavior balances exploration and exploitation in complex reasoning.


\begin{table}[!t]
\setlength\tabcolsep{4pt}
\renewcommand{\arraystretch}{0.9}
\small
\centering
    \begin{tabular*}{0.47\textwidth}{@{\extracolsep{\fill}}l cc@{}}
    \toprule
    \textbf{Method} & \textbf{GAIA} & \textbf{xbench} \\
    \midrule
    Qwen3-8B $\gets$ $\mathcal{E}$-\textit{8B / 32B} & 36.89 / 35.60 & 37.20 / 36.00 \\
    Qwen3-32B $\gets$ $\mathcal{E}$-\textit{32B / 8B} & 43.88 / 40.33 & 42.00 / 37.20 \\
    \bottomrule
    \end{tabular*}
    \caption{Experience repository swapping: $\mathcal{E}$-8B/32B denote dedicated repositories built for each agent.}
    \label{table:ana_trans}
\end{table}

\begin{figure*}[!t]  
\centering  
\includegraphics[width=16cm]{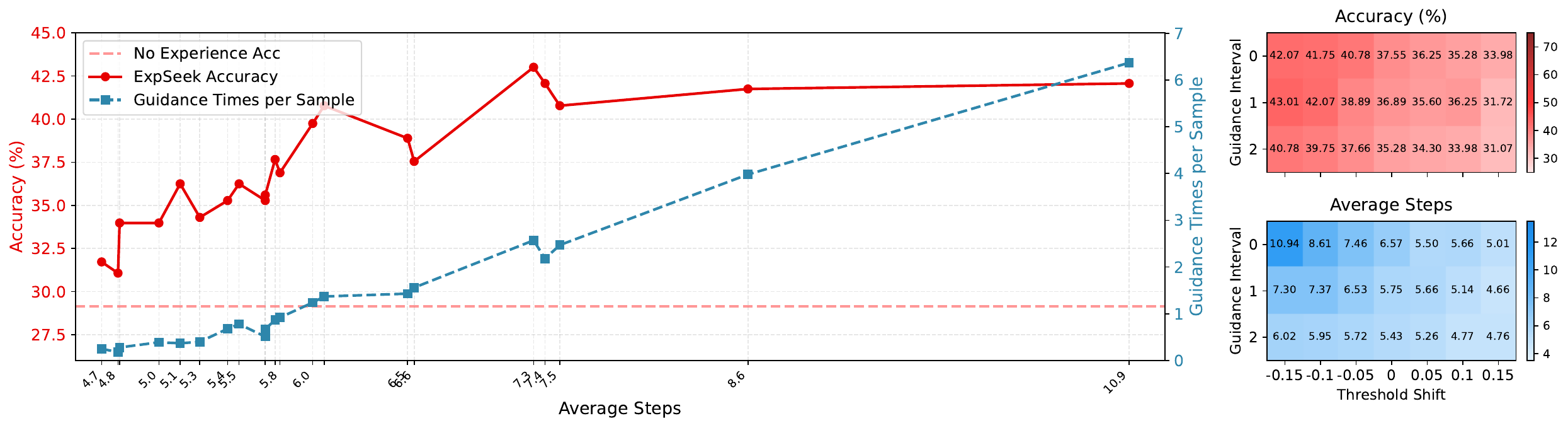}  
\caption{Cross-comparison results of performance and efficiency after adjusting intervention intensity.}
\label{fig:ana_acc_vs_eff}
\end{figure*}

\begin{figure}[!t]  
\centering  
\includegraphics[width=7.5cm]{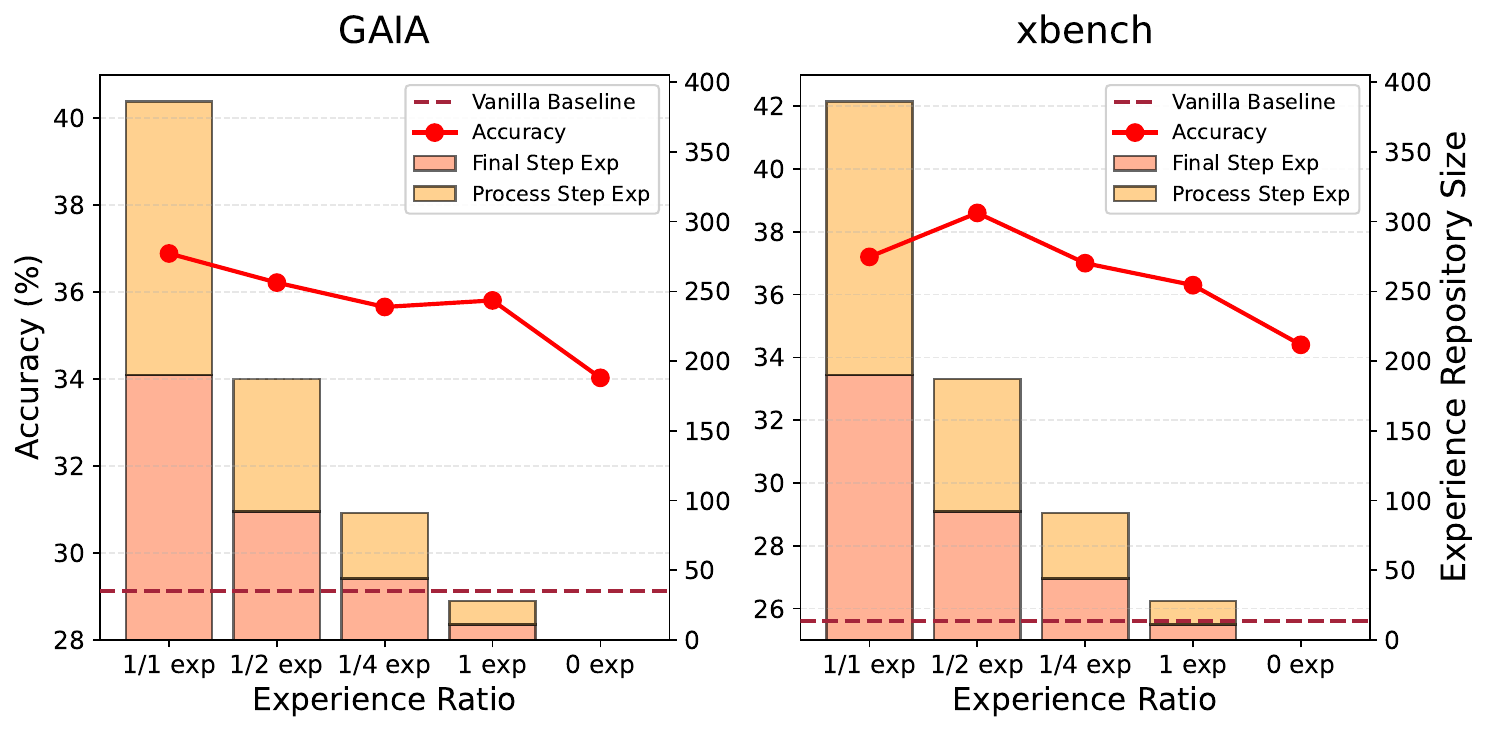}  
\caption{Correlation between repository size and performance on Qwen3-8B.}
\label{fig:ana_size}
\end{figure}

\subsection{Why Self-Trigger and Generation?}
To validate entropy-based self-triggering, we compare two variants: Reward Model-based\footnote{We employ \texttt{claude-sonnet-4-20250514} to judge intervention necessity at each step based on the full history.} (RM) and rule-based (continuous intervention from step one), both with one-step post-trigger silence.

Table~\ref{table:ana_claude4_vs_entropy} shows similar accuracy but divergent efficiency. Rule-based triggers result in 1.7$\times$ step and 2.6$\times$ time overhead on GAIA (1.5$\times$ and 2.1$\times$ on xbench) compared to entropy-based triggering. RM reduces step overhead (1.3-1.5$\times$) but increases time (2.2-2.9$\times$), suggesting over-intervention.
In contrast, self-trigger balances efficiency and performance by adapting intervention intensity to problem difficulty. On the more challenging xbench, its trigger rate increases 25.6\% compared to GAIA while maintaining similar accuracy, confirming its ability to precisely identify intervention timing based on internal state.

We also explore retrieval-based guidance, selecting the most similar experience (via \texttt{text-embedding-v4}\footnote{https://www.alibabacloud.com/help/en/model-studio/embedding}) instead of generation. Table~\ref{table:ana_claude4_vs_entropy} shows that retrieval-based guidance yields substantially lower accuracy than generative guidance, confirming the necessity of generative guidance.

\subsection{Does ExpSeek Scale and Transfer?}

We experiment with smaller 4B and 30B models\footnote{\texttt{Qwen3-4B-Instruct} and \texttt{Qwen3-30B-A3B-Instruct}} as $\mathcal{M}_e$ to explore the scaling of intervention models. As shown in Figure~\ref{fig:ana_small}, the three model sizes exhibit a scaling law on both GAIA and xbench, all achieving substantial performance gains. Notably, the 4B guidance model improves the 32B agent by 5.2\% and 9.7\% points respectively, validating the feasibility of weak-to-strong guidance given reasonable experience.
Additionally, we swap the experience pools between 8B and 32B Agents.
Results in Table~\ref{table:ana_trans} show that while experience has model dependency, the abstract guidance knowledge it contains still holds transfer value.

Furthermore, Figure~\ref{fig:ana_size} shows we reduce experiences per topic. Even with just one experience per topic, $\mathcal{M}_e$ still identifies key intervention points and maintains high accuracy, demonstrating that the experience model can understand and generalize to current reasoning even with only a few high-quality seed experiences, highlighting the necessity of experience topics. While removing the repository entirely degrades performance, the remaining substantial accuracy indicates that $\mathcal{M}_e$'s inherent world experience alone is beneficial.

\subsection{What Is the Efficiency Trade-off?}

Figure~\ref{fig:ana_acc_vs_eff} illustrates the relationship between accuracy and reasoning steps under varying intervention frequencies and thresholds. We expand the intervention interval from 1 to $\{0, 1, 2\}$ and shift the threshold by $\pm$ 0.05 three times, yielding 21 configurations. We conduct experiments with Qwen3-8B on the GAIA. Results show that as the trigger threshold decreases, reasoning steps increase rapidly while accuracy rises then plateaus with diminishing returns. With \textasciitilde2 interventions, accuracy reaches 43.01\%; beyond 6 interventions, performance barely improves, indicating that while increasing intervention intensity does not degrade performance, it also fails to yield higher gains. The results also demonstrate that in practice, the web agent produces stable performance even when thresholds fluctuate at a small scale.

\section{Conclusion}

We propose ExpSeek, a framework enabling web agents to actively seek step-level guidance based on step entropy. Even with small open-source agents, ExpSeek demonstrates significant performance improvements and excellent generalization on complex real-world web reasoning tasks, showing great potential for future development.

\section*{Limitations}

Although ExpSeek achieves significant performance advantages, it has the following limitations for future research: (1) The current threshold estimation relies on the training set and the tool model's assessment of step quality; more accurate strategies need to be investigated. (2) It remains unexplored whether ExpSeek can extend to non-web domains and integrate more tools. (3) Given ExpSeek's improvements in pass@k performance, it has not yet been studied whether it can serve as a rollout enhancement technique for Agentic Reinforcement Learning to improve convergence speed and sampling quality.

\section*{Ethical Considerations}
Our method is intended for academic research only and does not support applications involving risks, religion, racial discrimination, or ethical violations.

\section*{Acknowledgments}
We would like to thank the anonymous reviewers, the meta-reviewer, as well as the area chairs and program chairs for their valuable comments and efforts.
We would like to thank the members of IIE KDSec and Alibaba Tongyi Lab for their valuable discussions and exchanges.
This work is supported by the National Natural Science Foundation of China (Grant No. 62572465).

\normalem
\bibliography{custom}

\appendix

\section{Details of Method}
\label{sec:appendix}

\subsection{Details of Threshold Estimation}
\label{app:threshold}

We implement the threshold estimation algorithm using the \texttt{LogisticRegression} class from \texttt{scikit-learn} with default hyperparameters, which is well-suited for our single-feature binary classification task (correct vs. incorrect steps based on entropy $\bar{H}_t$).

\begin{algorithm}[!t]
\caption{Threshold Estimation}
\label{alg:threshold}
\begin{algorithmic}[1]
\INPUT Entropy sets $\mathcal{H}^+$ (correct steps), $\mathcal{H}^-$ (incorrect steps), Bootstrap iterations $N$

\OUTPUT Threshold interval $[\theta_{\text{lower}}, \theta_{\text{upper}}]$

\STATE Initialize $\Theta \gets \emptyset$

\FOR{$i = 1$ to $N$}
    \STATE \textcolor{gray}{\textit{// Bootstrap resampling}}
    \STATE Sample $\mathcal{H}^+_i$ by drawing $|\mathcal{S}^+|$ elements from $\mathcal{H}^+$ with replacement
    \STATE Sample $\mathcal{H}^-_i$ by drawing $|\mathcal{S}^-|$ elements from $\mathcal{H}^-$ with replacement
    
    \STATE \textcolor{gray}{\textit{// Construct training data}}
    \STATE $\mathcal{D}_i \gets \{(\bar{H}, 0) \mid \bar{H} \in \mathcal{H}^+_i\} \cup \{(\bar{H}, 1) \mid \bar{H} \in \mathcal{H}^-_i\}$
    
    \STATE \textcolor{gray}{\textit{// Train logistic regression}}
    \STATE Train logistic regression: $(w_i, b_i) \gets \mathrm{arg\,min}_{w,b} \sum_{(\bar{H},y) \in \mathcal{D}_i} \mathcal{L}(y, \sigma(w\bar{H}+b))$

    \STATE where $\sigma(z) = 1/(1+e^{-z})$
    
    \STATE \textcolor{gray}{\textit{// Extract threshold at decision boundary}}
    \STATE $\theta^{(i)} \gets -b_i / w_i$
    \STATE $\Theta \gets \Theta \cup \{\theta^{(i)}\}$
\ENDFOR

\STATE Sort $\Theta$ in ascending order
\STATE $\theta_{\text{lower}} \gets Q_{0.025}(\Theta)$
\STATE $\theta_{\text{upper}} \gets Q_{0.975}(\Theta)$

\RETURN $[\theta_{\text{lower}}, \theta_{\text{upper}}]$
\end{algorithmic}
\end{algorithm}

\paragraph{Bootstrap Resampling.}
Sampling with replacement (Lines 4-5 of Algorithm~\ref{alg:threshold}) generates bootstrap datasets where some original samples appear multiple times while others are omitted. Across $N=1000$ iterations, this produces a distribution of decision boundaries $\Theta = \{\theta^{(1)}, \ldots, \theta^{(N)}\}$, enabling uncertainty quantification via confidence intervals.

\paragraph{Decision Boundary Interpretation.}
Each logistic regression learns parameters $(w, b)$ such that the decision boundary $\theta = -b/w$ represents the entropy value where $P(\text{correct} \mid \bar{H}) = 0.5$. Geometrically, this corresponds to the inflection point of the sigmoid function $\sigma(w\bar{H}+b)$ projected onto the entropy axis—the point of maximum model uncertainty, making it a natural intervention threshold.

\paragraph{Confidence Interval Construction.}
The 95\% confidence interval $[\theta_{\text{lower}}, \theta_{\text{upper}}]$ is derived by extracting the 2.5th and 97.5th percentiles of the bootstrap distribution:
\begin{equation}
\theta_{\text{lower}} = Q_{0.025}(\Theta), \quad \theta_{\text{upper}} = Q_{0.975}(\Theta)
\end{equation}
where $Q_p(\cdot)$ denotes the $p$-th quantile function.

\paragraph{Computational Efficiency.}
On an Intel Xeon CPU, estimating thresholds for one step type (process or answer) completes within seconds. Since we estimate thresholds independently for both types, the total offline computation time is negligible compared to online inference costs.

\subsection{Additional Advantages of the Algorithm}
\begin{figure}[!t]  
\centering  
\includegraphics[width=7.7cm]{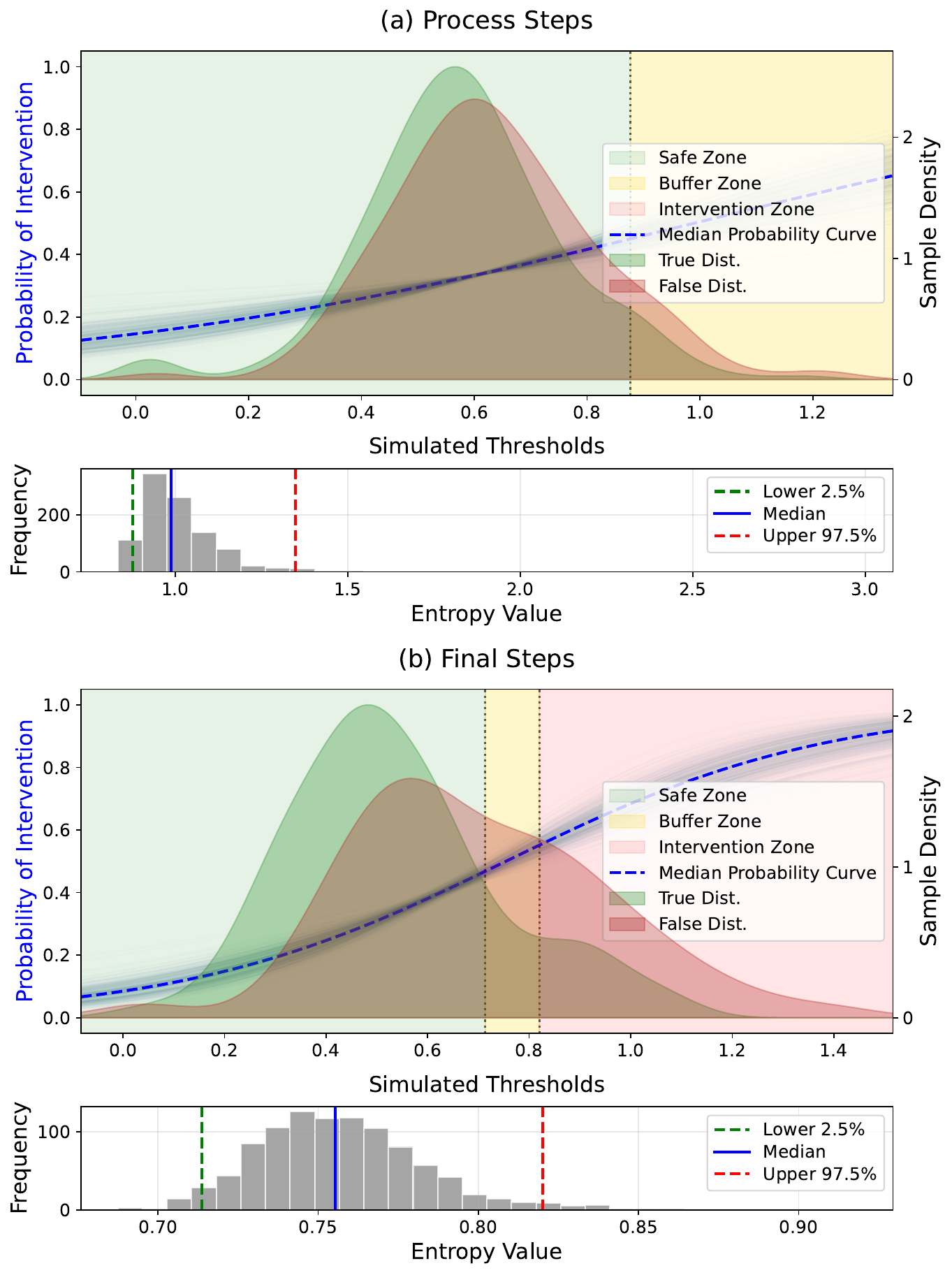}  
\caption{Entropy distributions of process and answer steps on $\mathcal{D}_{train}$ for Qwen3-32B, with fitted logistic regression curves.}
\label{fig:method_entropy_32b}
\end{figure}

As illustrated in Figure~\ref{fig:method_entropy_32b}, the threshold estimation algorithm adapts effectively to cases with reduced distributional differences. In the 32B process steps, the predominance of the yellow region indicates that under higher uncertainty, the algorithm adaptively randomizes trigger decisions to avoid excessive intervention.

\section{Details of Experiment Results}
\label{app:Experiment}

\begin{table*}[!ht]
\small
    \centering
    \begin{tabular*}{0.97\textwidth}{@{\extracolsep{\fill}}l cccc ccc c@{}}
    \toprule
    \multirow{2}{*}{\raisebox{-0.5\height}{\textbf{Method}}}
    & \multicolumn{4}{c}{\textbf{WebWalkerQA}} 
    & \multirow{2}{*}{\raisebox{-0.5\height}{\textbf{GAIA}}}
    & \multirow{2}{*}{\raisebox{-0.5\height}{\textbf{\textsc{Seal}}}} 
    & \multirow{2}{*}{\raisebox{-0.5\height}{\textbf{xbench}}}
    & \multirow{2}{*}{\raisebox{-0.5\height}{\textbf{Avg.}}} \\
    \cmidrule(lr){2-5}
    & \texttt{Easy} & \texttt{Medium} & \texttt{Hard} &  Avg. 
    & 
    & 
    & 
    & \\
    \midrule
    \multicolumn{9}{c}{\cellcolor{gray!15}\textit{Qwen3-8B}} \\

    No Experience & 57.50\textsubscript{\textcolor{gray}{- 0.0}} & 60.00\textsubscript{\textcolor{gray}{- 0.0}} & 46.11\textsubscript{\textcolor{gray}{- 0.0}} & 54.54\textsubscript{\textcolor{gray}{- 0.0}} & 47.57\textsubscript{\textcolor{gray}{- 0.0}} & 39.76\textsubscript{\textcolor{gray}{- 0.0}} & 43.00\textsubscript{\textcolor{gray}{- 0.0}} & 48.99\textsubscript{\textcolor{gray}{- 0.0}} \\
    
    Training-Free GRPO & 56.67\textsubscript{\textcolor{red}{$\downarrow$0.8}} & 62.54\textsubscript{\textcolor{blue}{$\uparrow$2.5}} & 52.67\textsubscript{\textcolor{blue}{$\uparrow$6.6}} & 57.29\textsubscript{\textcolor{blue}{$\uparrow$2.8}} & 48.54\textsubscript{\textcolor{blue}{$\uparrow$1.0}} & 42.13\textsubscript{\textcolor{blue}{$\uparrow$2.4}} & 45.00\textsubscript{\textcolor{blue}{$\uparrow$2.0}} & 51.26\textsubscript{\textcolor{blue}{$\uparrow$2.3}} \\
    
    \textsc{ReasoningBank}$^+$ & 61.17\textsubscript{\textcolor{blue}{$\uparrow$3.7}} & 58.19\textsubscript{\textcolor{red}{$\downarrow$1.8}} & 50.00\textsubscript{\textcolor{blue}{$\uparrow$3.9}} & 56.45\textsubscript{\textcolor{blue}{$\uparrow$1.9}} & 48.70\textsubscript{\textcolor{blue}{$\uparrow$1.1}} & 44.09\textsubscript{\textcolor{blue}{$\uparrow$4.3}} & 46.80\textsubscript{\textcolor{blue}{$\uparrow$3.8}} & 51.49\textsubscript{\textcolor{blue}{$\uparrow$2.5}} \\
    
    ExpSeek (Ours) & \textbf{69.17}\textsubscript{\textbf{\textcolor{blue}{$\uparrow$11.7}}} & \textbf{72.45}\textsubscript{\textbf{\textcolor{blue}{$\uparrow$12.5}}} & \textbf{63.33}\textsubscript{\textbf{\textcolor{blue}{$\uparrow$17.2}}} & \textbf{68.32}\textsubscript{\textbf{\textcolor{blue}{$\uparrow$13.8}}} & \textbf{54.37}\textsubscript{\textbf{\textcolor{blue}{$\uparrow$6.8}}} & \textbf{50.00}\textsubscript{\textbf{\textcolor{blue}{$\uparrow$10.2}}} & \textbf{62.00}\textsubscript{\textbf{\textcolor{blue}{$\uparrow$19.0}}} & \textbf{61.89}\textsubscript{\textbf{\textcolor{blue}{$\uparrow$12.9}}} \\

    \midrule
    \multicolumn{9}{c}{\cellcolor{gray!15}\textit{Qwen3-32B}} \\
    

    No Experience & 62.50\textsubscript{\textcolor{gray}{- 0.0}} & 61.43\textsubscript{\textcolor{gray}{- 0.0}} & 58.33\textsubscript{\textcolor{gray}{- 0.0}} & 60.75\textsubscript{\textcolor{gray}{- 0.0}} & 54.37\textsubscript{\textcolor{gray}{- 0.0}} & 47.64\textsubscript{\textcolor{gray}{- 0.0}} & 53.00\textsubscript{\textcolor{gray}{- 0.0}} & 56.21\textsubscript{\textcolor{gray}{- 0.0}} \\
    
    Training-Free GRPO & 63.95\textsubscript{\textcolor{blue}{$\uparrow$1.5}} & 59.57\textsubscript{\textcolor{red}{$\downarrow$1.9}} & 57.45\textsubscript{\textcolor{red}{$\downarrow$0.9}} & 60.32\textsubscript{\textcolor{red}{$\downarrow$0.4}} & 56.62\textsubscript{\textcolor{blue}{$\uparrow$2.3}} & 48.35\textsubscript{\textcolor{blue}{$\uparrow$0.7}} & 53.60\textsubscript{\textcolor{blue}{$\uparrow$0.6}} & 56.59\textsubscript{\textcolor{blue}{$\uparrow$0.4}} \\
    
    \textsc{ReasoningBank}$^+$ & 68.33\textsubscript{\textcolor{blue}{$\uparrow$5.8}} & 64.33\textsubscript{\textcolor{blue}{$\uparrow$2.9}} & 56.44\textsubscript{\textcolor{red}{$\downarrow$1.9}} & 63.03\textsubscript{\textcolor{blue}{$\uparrow$2.3}} & 46.60\textsubscript{\textcolor{red}{$\downarrow$7.8}} & 48.82\textsubscript{\textcolor{blue}{$\uparrow$1.2}} & 59.00\textsubscript{\textcolor{blue}{$\uparrow$6.0}} & 57.25\textsubscript{\textcolor{blue}{$\uparrow$1.0}} \\
    
    ExpSeek (Ours) & \textbf{72.50}\textsubscript{\textbf{\textcolor{blue}{$\uparrow$10.0}}} & \textbf{72.86}\textsubscript{\textbf{\textcolor{blue}{$\uparrow$11.4}}} & \textbf{63.33}\textsubscript{\textbf{\textcolor{blue}{$\uparrow$5.0}}} & \textbf{69.56}\textsubscript{\textbf{\textcolor{blue}{$\uparrow$8.8}}} & \textbf{63.11}\textsubscript{\textbf{\textcolor{blue}{$\uparrow$8.7}}} & \textbf{51.97}\textsubscript{\textbf{\textcolor{blue}{$\uparrow$4.3}}} & \textbf{66.00}\textsubscript{\textbf{\textcolor{blue}{$\uparrow$13.0}}} & \textbf{64.96}\textsubscript{\textbf{\textcolor{blue}{$\uparrow$8.8}}} \\

    \bottomrule
    \end{tabular*}
    \caption{Extended main results reporting Pass@3 accuracy for baselines and ExpSeek. We also provide absolute performance differences from experience-free ReAct.}
    \label{table:main_exp_pass3}
\end{table*}

\subsection{Pass@k Results}
As shown in Figure~\ref{table:main_exp_pass3}, ExpSeek demonstrates stronger performance in sampling diversity. Compared to vanilla, the absolute improvements in pass@3 are 12.9\% and 8.8\%, exceeding the improvement margins in average accuracy. In contrast, the two baselines do not exhibit better diversity, and even fall below vanilla on multiple datasets. This highlights the significant advantage of ExpSeek, with potential even as a rollout augmentation strategy in agentic RL training.

\subsection{Why prior methods underperform?}

Based on the experimental results, we reflect on why prior experience-based intervention methods are ineffective. We attribute the reasons to two aspects.

(1) \textit{Experience acquisition is disconnected from actual reasoning}. Most experience repository construction methods meticulously process training trajectories, perform multiple rounds of careful denoising, and cautiously distill core experience items with deep observation of successes and failures. However, the construction process itself is challenging, and when such heavily processed experiences are given to reasoning agents, the agents may not even understand some abstract expressions. This differs from how humans use experience—when humans recall experience, the amount of information retrieved instantly is enormous, including scenes, behavioral details, and even emotions, far exceeding simplified experience items.

(2) \textit{Experience is difficult to utilize}. During multi-turn agent interactions, even when accurate experiences are provided in system prompts, it is difficult to require models to precisely locate a few short effective experience items across ultra-long contexts. Moreover, models must also leverage experience to correct their original reasoning tendencies.

ExpSeek essentially simulates a user through an experience model, providing effective guidance that does not require deep model understanding to be leveraged. Since each guidance is generated in real-time based on context, the experience is not disconnected from actual reasoning.

\begin{figure*}[!t] 
\centering  
\includegraphics[width=16cm]{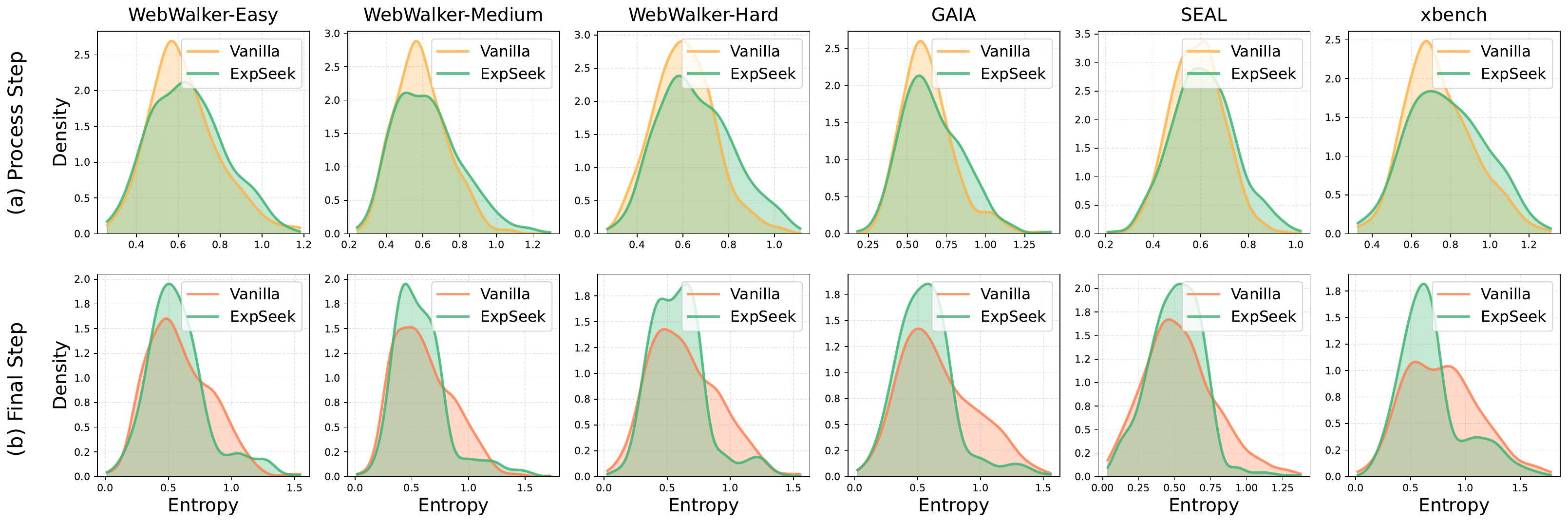}
\caption{Entropy distributions of process and answer steps for Qwen3-32B before and after applying ExpSeek across all benchmarks.}
\label{fig:qwen3_32b_six}
\end{figure*}

\subsection{Case Studies}

As shown in Table~\ref{table:case_incorrect1},~\ref{table:case_incorrect2},~\ref{table:case_correct1}and~\ref{table:case_correct2}.
The contrasting outcomes reveal a critical limitation in agent's reasoning when relying on incomplete information, and demonstrate how strategic guidance can redirect agents toward reliable evidence.

\textbf{Why the unguided trajectory failed}: The agent's error stemmed from reliance on search snippets that listed "Spider-Man: No Way Home" alongside "The Super Mario Bros. Movie" and "Jurassic World: Dominion" in billion-dollar film discussions, but omitted distributor information. Without verification mechanisms, the agent incorrectly inferred all three were Universal releases, when "Spider-Man: No Way Home" was actually distributed by Sony Pictures. This represents a classic entity attribute confusion—the agent failed to verify the critical "distributor" attribute before counting.

\textbf{Why guidance succeeded}: The guidance implemented a two-pronged strategy. First, it redirected the agent toward authoritative box office sites (Screen Rant, Rotten Tomatoes, Box Office Mojo) that systematically label films with distributors. Second, it reinforced the task's dual requirement—counting films satisfying both revenue threshold and distributor constraint—prompting a verification-based approach. The Screen Rant article provided explicit evidence that only two Universal-distributed films (via Illumination and Amblin partnerships) appeared on the billion-dollar list, enabling correct filtering and counting.

The guidance's success lies in methodological redirection rather than direct answer provision. By steering the agent toward sources with richer structural metadata and emphasizing attribute verification, it enabled the agent to overcome reasoning failure through improved evidence quality rather than external correction.

\begin{table*}[!t]
\renewcommand{\arraystretch}{1.0}
\small
\centering
    \begin{tabularx}{1.0\textwidth}{X}
    \toprule
    \textbf{Experience base demos.} \\
    \midrule
    \textit{Process Step demo} \\
    \midrule
    \texttt{\{} \\
    \quad "\textbf{Behavior}": "The student recognized the need to confirm which African countries participated in the 7th China International Import Expo (CIIE) under the theme of '60th Anniversary of Diplomatic Relations' through external information sources, and therefore initiated multi-angle keyword searches.", \\
    \\
    
    \quad "\textbf{Mistake}": "Although the search keywords were reasonably designed, the student did not prioritize visiting sources most likely to contain authoritative information (such as the official CIIE website) for in-depth access. Instead, they relied only on search result snippets for preliminary judgment, laying hidden risks for subsequent misjudgment.", \\
    \\
    
    \quad "\textbf{Guidance}": "When reviewing search results, you should prioritize identifying and planning to visit authoritative or official sources (such as ciie.org). Especially when multiple results cite the same webpage, it should be set as the primary verification target to avoid prematurely synthesizing incomplete information to draw conclusions." \\
    
    \texttt{\}}  \\

    \midrule
    \textit{Final Step demo} \\
    \midrule
    \texttt{\{} \\
    \quad "\textbf{Behavior}": "The student reviewed the search results, found that multiple results pointed to a website design company called 'Niegao' (niegoweb.com), and inferred from this that the website was designed by this company.", \\
    \\
    \quad "\textbf{Mistake}": "The student overlooked the complete official company name that clearly appeared in the search results—'Niegao Brand Network' or 'Niegao Brand Design'—and instead simplified it on their own to 'Niegao Website Design Company,' resulting in an answer inconsistent with the standard answer. Although semantically similar, the student failed to faithfully reproduce the proper noun from the original text.",\\
    \\
    \quad "\textbf{Guidance}": "Before drawing conclusions, please carefully check the complete institutional names that appear in the search results, especially the official names of collaborating parties in case study pages. Prioritize using the full brand name directly mentioned in the original text rather than summarizing it based on your own understanding."\\

    \texttt{\}}  \\
    \bottomrule
    \end{tabularx}
    \caption{Experience base demos for Qwen3-8B.}
    \label{table:exp_base}
\end{table*}



\subsection{Guiding Experience}
In this section, we demonstrate the details of the guiding experience. Table~\ref{table:exp_base} presents experience demos from both stages. The process step experience focuses on strategic decisions during reasoning, such as how the student selects information sources for verification after searching. Its mistake identifies methodological issues affecting subsequent reasoning paths (e.g., relying on snippets rather than accessing authoritative sources), while guidance provides directional suggestions for process optimization (e.g., prioritizing official sources). In contrast, the final step experience concentrates on answer formulation, such as extracting the final answer from search results. Its mistake targets detail-oriented errors directly impacting output quality (e.g., failing to faithfully reproduce complete proper nouns), while guidance emphasizes result accuracy, prompting the model to focus on answer details and encouraging trajectory extension for thorough verification when necessary. This triplet design reproduces step-level error patterns from training trajectories and provides targeted guidance, enabling the experience model to more accurately guide both process and final steps during test inference.

\section{Details of Setting}
\label{app:Setting}
Following the LLM-as-a-judge approach~\cite{zheng2023judging,zhang2023widerdeeperllmnetworks,li-etal-2025-generation}, we use \texttt{Qwen3-235B-A22B-Instruct-2507} as the evaluator with temperature set to 0 and top-p set to 1 (see Table~\ref{table:prompt_eval} for the evaluation prompt).
As Qwen3-8B and Qwen3-32B are hybrid reasoning models, we deploy them in non-thinking mode and adopt the <thought></thought> tag rather than <think></think> to prevent conflicts with their original output structure.
For tool implementation, we use \textit{Bright Data}\footnote{https://www.bright.cn} to provide stable web API services for the search tool, \textit{Jina}\footnote{https://jina.ai} as the web access service for the visit tool, and \texttt{Qwen3-235B-A22B-Instruct-2507} as the summarization model within the visit tool. All other experimental settings and prompts remain consistent with prior work~\cite{wu2025webdancer}.
The four benchmarks employed in our evaluation are widely recognized in web agent research and are permitted for use in academic studies under their distribution terms.

\section{Extended Related Work}
\label{app:extended_rw}

The related work in the main text is organized along two technical dimensions: experience intervention and entropy.
This section introduces the broader background and recent progress in agent research.

Agents in long-horizon interactions rely on effective memory mechanisms to maintain coherence.
Recent studies have advanced long-term memory through hierarchical and graph-based structuring~\cite{yue2026hypermem,wu2026GAM}, temporal-semantic modeling~\cite{su2026dialogue}, episodic representations~\cite{ma2026cast}, dynamic context management in non-linear dialogues~\cite{hu2026context}, and autonomous memory compression~\cite{li2026mempo}.
Complementarily, challenges of redundancy, noise, and ordering sensitivity in retrieved information have been addressed via adaptive compression, minimal sufficient clue selection, and permutation robustness~\cite{zhang2024adacomp,compselect,zhang2026stable}, while IP protection for retrieval-augmented systems~\cite{liu2025stole} and multi-agent collaborative fact verification~\cite{wu2025zofia} have also attracted attention.

The ability of agents to continuously self-improve is another active direction.
A line of work advances agent self-evolution through trajectory-level optimization~\cite{lin2025se}, experience-driven agent creation~\cite{hao2026recreate}, co-evolutionary capability expansion and experience distillation~\cite{cheng2026mem2evolve}, curriculum-based feedback alignment~\cite{li2025curriculum}, and test-time variational synthesis over unlabeled queries~\cite{bai2026ttvs}.
For tool use, closed-loop data evolution that couples capability diagnosis with targeted data refinement has proven highly effective~\cite{zhang2025looptool}.
At the policy level, adaptive reasoning depth switching~\cite{wang2026adaptive}, turn-level reward assignment in multi-turn search~\cite{ma2026tspo}, and efficient multi-agent communication topology generation~\cite{sun2026topodim} have improved decision-making efficiency, and a theoretical analysis of token-level entropy dynamics during reinforcement learning~\cite{hao2025rethinking} provides an important reference for understanding entropy's role in agent systems.

Agent research also continues to expand toward broader modalities and scenarios.
In multimodal and embodied domains, visual-textual retrieval evaluation~\cite{zeng2026vision}, self-supervised multimodal model refinement~\cite{han2026unicorn}, self-critique under tool-calling errors~\cite{huang2025critictool}, partial decoupling of reasoning and action~\cite{fang2025dualvla}, cross-embodied foundation models~\cite{hao2025mimo}, UI-in-the-Loop GUI reasoning~\cite{li2026whatsmissing}, unified spatial intelligence modeling with auxiliary modality generation~\cite{zhang2025cooper}, and explainable multimodal disaster assessment~\cite{ZHANG2026112674} collectively advance perception and execution capabilities.
Research on composed video and image retrieval~\cite{ReTrack,REFINE,HABIT,liu2025queries} further improves the precision of multimodal information acquisition.
In social intelligence, strategy-injected agent training~\cite{zhang-etal-2025-sotopia}, role-playing agents~\cite{zhang-etal-2025-revealing,chen2026actormindemulatinghumanactor}, collaborative reasoning under imperfect-information games~\cite{zhong2026collaborative}, large-scale social simulation integrating macro-micro modeling~\cite{zhang2026coupling,zhang-etal-2025-ga}, and multi-agent negotiation~\cite{long2026emomas} provide new paradigms for studying agent behavior in open environments.
Reasoning over noisy structured data~\cite{ye2025tableqa,guo2026rethinking} and knowledge retention in continual learning~\cite{pcgr2025naacl,he-etal-2025-dont} also remain active directions.
Collectively, these diverse lines reflect a broader trend of agents shifting from passive execution toward proactive decision-making in complex environments, and the intrinsic-signal-driven experience seeking explored by ExpSeek represents a concrete manifestation of this trend in the web agent setting.

\section{Details of Prompts}
\label{app:prompt_detail}
In this section, we present all key prompts.
Table~\ref{table:prompt_react} shows the system prompt for ReAct, which we have slightly adjusted based on guidance experience. Tables~\ref{table:prompt_generate_triplet} and~\ref{table:prompt_generate_topic} are prompts for generating the experience base, where Table~\ref{table:prompt_generate_triplet} implements identifying step mistakes and generating triplets, and Table~\ref{table:prompt_generate_topic} iteratively generates topics based on all triplets.
Tables~\ref{table:prompt_stage_one} and~\ref{table:prompt_stage_two} are prompts for utilizing the experience model during inference. Table~\ref{table:prompt_stage_one} is for the experience model to determine which topics to utilize, and Table~\ref{table:prompt_stage_two} generates appropriate proactive guidance based on all experiences under the topics and the current history context.
Table~\ref{table:prompt_model_as_trigger} is the prompt for using a LLM as an intervention trigger, where outputting ``yes'' triggers guidance for that step. Table~\ref{table:prompt_emb} is for utilizing retrieved experience base experiences during inference. Table~\ref{table:prompt_direct_guide} is for generating guidance directly using the world experience of the experience model without relying on the experience base.

\begin{table*}[t]
\renewcommand{\arraystretch}{1.0}
\small
\centering
    \begin{tabularx}{\textwidth}{X}
    \toprule
    \textbf{Case study (Incorrect): Qwen3-8B interaction without guiding experience} \\
    \midrule
    \textbf{Question:} \\
    How many Universal Studios films have grossed over \$1 billion worldwide since 2020? \\

    \\
    
    \textbf{Agent Action:} \\
    <think>\\
    To determine how many Universal Studios films have grossed over \$1 billion worldwide since 2020, I need to search for a list of Universal Studios films and their box office earnings. This will involve first identifying which films are associated with Universal Studios and then checking their gross earnings. I'll begin by searching for a list of Universal Studios films and their box office performance since 2020.\\
    </think> \\
    <tool\_call>\\
    Search: Universal Studios films gross over \$1 billion since 2020\\
    Search: List of Universal Studios movies with box office earnings over \$1 billion since 2020\\
    </tool\_call>\\
    
    \textbf{Env Tool Response:} \\
    A Google search for 'Universal Studios films gross over \$1 billion since 2020' found 5 results:\\
    ......\\
    2. \lbrack Universal - The First Half of the 2020s In Review : r/boxoffice\rbrack\\
       Universal - The First Half of the 2020s In Review · The Super Mario Bros. Movie - \$1.36B · Jurassic World Dominion - \$1.001B · Oppenheimer- \$975.6M.\\
    ......\\
    5. \lbrack6 Movies That Have Made \$1 Billion At The Box Office Since Covid\rbrack\\
       6 Jurassic World: Dominion (2022) · 5 The Super Mario Bros. Movie (2023) · 4 Spider-Man: No Way Home (2021) · 3 Avatar: The Way of Water (2022) · 2 ...\\
    ......\\
    \\
    \textbf{Agent Action:}\\
    <think>\\
    Based on the search results, I need to determine which Universal Studios films have grossed over \$1 billion worldwide since 2020. The key information comes from the Reddit post and the The Numbers website. According to the Reddit post, "The Super Mario Bros. Movie" has grossed \$1.36B, and "Jurassic World Dominion" has grossed \$1.001B. From the The Numbers website, there is a list of the top Universal movies by worldwide box office, and the most recent ones include "The Super Mario Bros. Movie" and "Jurassic World Dominion." Additionally, it appears that other films such as "Avatar: The Way of Water" might also qualify based on the context provided in the search results. To confirm this, I will visit the The Numbers website directly to get the exact data for Universal Studios films and their box office earnings since 2020.\\
    </think>\\
    <tool\_call>\\
    Visit: https://www.the-numbers.com/box-office-records/worldwide/all-movies/theatrical-distributors/universal\\
    Visit Goal: Find the list of Universal Studios films with box office earnings over \$1 billion since 2020.\\
    </tool\_call>\\
    \\
    \textbf{Env Tool Response:} \\
    The requested webpage did not load fully and contains no usable data. As a result, there is insufficient information to identify Universal Studios films with box office earnings exceeding \$1 billion since 2020.\\
    ......\\
    \\    
    \textbf{Agent Action:} \\
    <think> \\
    The attempt to access the The Numbers website directly did not yield usable data. This may be due to technical issues or incomplete page loading. I will need to find an alternative source of information. Based on the previous search results, the Reddit post mentioned "The Super Mario Bros. Movie" and "Jurassic World Dominion" as having grossed over \$1 billion. However, I should verify this with another reliable source to confirm the accuracy of these figures.\\
    </think> \\
    <tool\_call>\\
    Search: Universal Studios films over \$1 billion since 2020\\
    Search: List of Universal Studios movies with box office earnings over \$1 billion since 2020\\
    </tool\_call>\\
    
    \bottomrule
    \end{tabularx}
    \caption{Case study: Qwen3-8B interaction without guiding experience}
    \label{table:case_incorrect1}
\end{table*}

\begin{table*}[t]
\renewcommand{\arraystretch}{1.0}
\small
\centering
    \begin{tabularx}{\textwidth}{X}
    \toprule
    \textbf{Case study (Incorrect): Qwen3-8B interaction without guiding experience} \\
    \midrule
    
    \textbf{Env Tool Response:} \\
    A Google search for 'Universal Studios films over \$1 billion since 2020' found 5 results:\\
    ......\\
    3. \lbrack8 Movies That Broke \$1 Billion at the Box Office Since Covid\rbrack
       Avatar: The Way of Water's box office comes in at over \$2.3 billion...\\
    ......\\
    5. \lbrack6 Movies That Have Made \$1 Billion At The Box Office Since Covid\rbrack
       6 Jurassic World: Dominion (2022) · 5 The Super Mario Bros. Movie (2023) · 4 Spider-Man: No Way Home (2021) · 3 Avatar: The Way of Water (2022) · 2 ...\\
    ......\\
    \\
    \textbf{Agent Action:} \\
    <think> \\
    From the search results, I can see that "The Super Mario Bros. Movie" (2023) and "Jurassic World: Dominion" (2022) are mentioned as having grossed over \$1 billion worldwide. Additionally, "Spider-Man: No Way Home" (2021) is noted as one of the first films to reach the \$1 billion mark since the start of the pandemic. These three films are distributed by Universal Studios, and they all have box office earnings exceeding \$1 billion globally since 2020. Therefore, the answer to the question is that three Universal Studios films have grossed over \$1 billion worldwide since 2020. \\
    </think> \\
    <answer> \\
    Three Universal Studios films have grossed over \$1 billion worldwide since 2020: "The Super Mario Bros. Movie," "Jurassic World: Dominion," and "Spider-Man: No Way Home." \\
    </answer> \\
    
    \bottomrule
    \end{tabularx}
    \caption{Case study: Qwen3-8B interaction without guiding experience (Table continued)}
    \label{table:case_incorrect2}
\end{table*}

\begin{table*}[t]
\renewcommand{\arraystretch}{1.0}
\small
\centering
    \begin{tabularx}{\textwidth}{X}
    \toprule
    \textbf{Case study (Correct): Qwen3-8B interaction with guiding experience} \\
    \midrule

    \textbf{Question:}\\
    How many Universal Studios films have grossed over \$1 billion worldwide since 2020? \\
    \\
    \textbf{Agent Action:}\\
    <think> \\
    To determine how many Universal Studios films have grossed over \$1 billion worldwide since 2020, I need to search for a list of Universal Studios films and their box office earnings. This will involve first identifying which films are associated with Universal Studios and then checking their gross earnings. I'll begin by searching for a list of Universal Studios films and their box office performance since 2020. \\
    </think> \\
    <tool\_call> \\
    Search: Universal Studios films gross over \$1 billion since 2020\\
    Search: List of Universal Studios movies with box office earnings over \$1 billion since 2020\\
    </tool\_call> \\
     \\
    \textbf{Env Tool Response:} \\
    A Google search for 'Universal Studios films gross over \$1 billion since 2020' found 5 results: \\
    ...... \\
    2. \lbrack Universal - The First Half of the 2020s In Review : r/boxoffice\rbrack \\
       Universal - The First Half of the 2020s In Review · The Super Mario Bros. Movie - \$1.36B · Jurassic World Dominion - \$1.001B · Oppenheimer- \$975.6M. \\
    ...... \\
    5. \lbrack6 Movies That Have Made \$1 Billion At The Box Office Since Covid\rbrack \\
       6 Jurassic World: Dominion (2022) · 5 The Super Mario Bros. Movie (2023) · 4 Spider-Man: No Way Home (2021) · 3 Avatar: The Way of Water (2022) · 2 ... \\
    ...... \\
     \\
    \textbf{Agent Action:} \\
    <think> \\
    Based on the search results, I need to determine which Universal Studios films have grossed over \$1 billion worldwide since 2020. The key information comes from the Reddit post and the The Numbers website. According to the Reddit post, "The Super Mario Bros. Movie" has grossed \$1.36B, and "Jurassic World Dominion" has grossed \$1.001B. From the The Numbers website, there is a list of the top Universal movies by worldwide box office, and the most recent ones include "The Super Mario Bros. Movie" and "Jurassic World Dominion." Additionally, it appears that other films such as "Avatar: The Way of Water" might also qualify based on the context provided in the search results. To confirm this, I will visit the The Numbers website directly to get the exact data for Universal Studios films and their box office earnings since 2020. \\
    </think> \\
    <tool\_call> \\
    Visit: https://www.the-numbers.com/box-office-records/worldwide/all-movies/theatrical-distributors/universal\\
    Visit goal: Find the list of Universal Studios films with box office earnings over \$1 billion since 2020\\
    </tool\_call> \\
     \\
    \textbf{Env Tool Response:}  \\
    The requested webpage did not load fully and contains no usable data. As a result, there is insufficient information to identify Universal Studios films with box office earnings exceeding \$1 billion since 2020. \\
    ...... \\
     \\
    \textbf{Agent Action:} \\
    <think> \\
    The attempt to access the The Numbers website directly did not yield usable data. This may be due to technical issues or incomplete page loading. I will need to find an alternative source of information. Based on the previous search results, the Reddit post mentioned "The Super Mario Bros. Movie" and "Jurassic World Dominion" as having grossed over \$1 billion. However, I should verify this with another reliable source to confirm the accuracy of these figures. \\
    </think> \\
    <tool\_call> \\
    Search: Universal Studios films over \$1 billion since 2020\\
    Search: List of Universal Studios movies with box office earnings over \$1 billion since 2020\\
    </tool\_call> \\

    \bottomrule
    \end{tabularx}
    \caption{Case study: Qwen3-8B interaction with guiding experience}
    \label{table:case_correct1}
\end{table*}

\begin{table*}[t]
\renewcommand{\arraystretch}{1.0}
\small
\centering
    \begin{tabularx}{\textwidth}{X}
    \toprule
    \textbf{Case study (Correct): Qwen3-8B interaction with guiding experience} \\
    \midrule
    
    \textbf{Env Tool Response (with Guidance):} \\
    A Google search for 'Universal Studios films over \$1 billion since 2020' found 5 results: \\
    ...... \\
    3. \lbrack8 Movies That Broke \$1 Billion at the Box Office Since Covid\rbrack \\
       Avatar: The Way of Water's box office comes in at over \$2.3 billion... \\
    ...... \\
    5. \lbrack6 Movies That Have Made \$1 Billion At The Box Office Since Covid\rbrack \\
    6 Jurassic World: Dominion (2022) · 5 The Super Mario Bros. Movie (2023) · 4 Spider-Man: No Way Home (2021) · 3 Avatar: The Way of Water (2022) · 2 ... \\
    ...... \\
    <guidance>
    Review the core of the question: you need to count films released by Universal Studios with worldwide box office over \$1 billion since 2020. This means the answer is not listing a specific film, but a complete count based on reliable sources.
    You have found some films that may meet the criteria, but different sources show inconsistencies. Instead of repeating previous searches, carefully examine those webpages in the existing search results that specifically compile "billion-dollar movie" lists, such as links from Screen Rant, Rotten Tomatoes, or Box Office Mojo. These pages often label each film with its distributor and specific box office figures, helping you verify one by one whether it belongs to Universal Studios and meets the time and amount conditions.
    Special reminder: Even if a page doesn't directly state "there are X films in total," you can still obtain the answer through manual counting—the key is finding a reliable list that completely catalogs such films and marks their distributors. Prioritize visiting such summary articles or professional box office ranking pages.
    </guidance> \\
    \\
    \textbf{Agent Action: }\\
    <think> \\
    Based on the guidance, I should focus on reliable sources that list billion-dollar movies and identify which ones are distributed by Universal Studios. The Screen Rant article mentioned "Jurassic World: Dominion" and "The Super Mario Bros. Movie" as having grossed over \$1 billion, but I need to confirm this with direct access to the full article or a more comprehensive list. To do this, I will visit the Screen Rant page to verify the information. \\
    </think> \\
    <tool\_call> \\
    Visit: https://screenrant.com/billion-dollar-movies-box-office-2020\\
    Visit goal: Identify the number of Universal Studios films with box office earnings over \$1 billion since 2020\\
    </tool\_call> \\
    \\
    \textbf{Env Tool Response:} \\
    Evidence in page: \\
    \lbrack"5 The Super Mario Bros. Movie (2023) - The movie's success has led to Nintendo teasing a range of films based on more of its iconic video game characters.", "Illumination earned a much higher net profit than other billion-dollar movies", "6 Jurassic World: Dominion (2022) - However, after scathing reviews... the movie was always going to flounder. The movie did face unlikely box office competition, too. Top Gun: Maverick was still playing when Jurassic World: Dominion was released...", "Jurassic World: Dominion made almost exactly \$1 billion worldwide."\rbrack \\
     \\
    Summary: \\
    Since 2020, two Universal Studios films have earned over \$1 billion at the global box office: *Jurassic World: Dominion* (2022) and *The Super Mario Bros. Movie* (2023). Both were produced by Universal-owned or partnered studios—Illumination for *The Super Mario Bros. Movie* and Amblin Entertainment in partnership with Universal for *Jurassic World: Dominion*. These are the only two Universal-released films on the list of six post-2020 billion-dollar earners. \\
     \\
    \textbf{Agent Action:} \\
    <think> \\
    Based on the information from the Screen Rant article, there are exactly two Universal Studios films that have grossed over \$1 billion worldwide since 2020: "Jurassic World: Dominion" (2022) and "The Super Mario Bros. Movie" (2023). These are the only two films from Universal Studios listed among the billion-dollar earners since 2020. \\
    </think> \\
    <answer> \\
    Since 2020, two Universal Studios films have grossed over \$1 billion worldwide: "Jurassic World: Dominion" (2022) and "The Super Mario Bros. Movie" (2023). Therefore, the answer is 2. \\
    </answer> \\

    \bottomrule
    \end{tabularx}
    \caption{Case study: Qwen3-8B interaction with guiding experience  (Table continued)}
    \label{table:case_correct2}
\end{table*}

\begin{table*}[ht]
\renewcommand{\arraystretch}{1.0}
\small
\centering
    \begin{tabularx}{\textwidth}{X}
    \toprule
    \textbf{ReAct system prompt.} \\
    \midrule

    You are a problem-solving expert, skilled at analyzing complex problems, forming solutions, and gradually resolving them step-by-step. Your approach is flexible and can be adjusted based on the progress of problem-solving. Any problem you encounter will certainly be solvable; therefore, no matter how complex, you must not give up until you find the answer you consider correct.   \\
     \\
    You must output two parts based on the current input, both enclosed by specific angle bracket position markers:   \\
    **The first part** is enclosed by <thought></thought> and contains your self-thinking process, including analysis of the problem, determining whether tool usage is required, which tool to call, and what values to pass to the tool.   \\
    **The second part** is divided into two situations:   \\
        \quad(1) If you wish to call a tool, then after </thought> output <tool\_call></tool\_call> containing standard tool call content, with the format introduced below.   \\
        \quad(2) If you believe the current problem can already be solved, then after </thought> output <answer></answer> containing the final answer.   \\
     \\
    **Available tools and their function overview** \\
    1. search: Input a keyword you think needs to be searched, returns multiple website links and corresponding summaries.   \\
    2. visit: Input a website link and the problem you hope to solve by visiting that site, returns the problem's answer and a snippet of the original content from the site corresponding to the answer.   \\
     \\
    **Detailed tool call format**   \\
    \{search\_format\} \\
    \{visit\_format\} \\
     \\
    **Model workflow description**   \\
    1. Model input consists of the historical loop of interactions, including the model’s generated thoughts and tool calls, and the user's returned tool results. The history may have multiple rounds, all designed to solve a given problem.  \\ 
    2. The model must output two parts: thinking process, and tool call or answer, each enclosed by specific angle bracket position markers:   \\
        \quad- If thinking, follow the format: <thought> here is the thinking process </thought>   \\
        \quad- If calling a tool, follow the format: <tool\_call> tool call here </tool\_call>   \\
        \quad- If determining the final answer, follow the format: <answer> final answer </answer>   \\
    3. The user will respond with tool call results or occasionally provide guidance. If there is guidance, you should carefully consider whether the user's ideas are reasonable and try to follow them: \\
        \quad- If responding to a tool call, you will see the format: <tool\_response> here is the tool's return value </tool\_response> \\
        \quad- If the user provides careful guidance, you will see the format: <user\_guidance> here is the user's guidance content </user\_guidance> \\
    4. Every problem must have an answer; during multi-round resolution, do not forget your past planning and process results, and do not forget the details embedded in the problem.   \\
    5. The search tool's return value is only website links and snippet summaries; they are hardly reliable references and can only serve as search direction. To obtain accurate information, you must call the visit tool to visit a site.   \\
    6. If the interaction includes a resolution plan, follow the plan. Do not blindly ignore key constraints in the plan to avoid potential cascading errors.   \\
    7. In tool\_call generation, the format must follow the above definitions and be valid JSON. An incorrect format will cause tool calls to fail.   \\
    *8*. **IMPORTANT**: If the user provides guidance after the answer, prioritize regenerating <tool\_call></tool\_call> to continue searching for missing clues, or provide only when you are absolutely certain of the answer. \\
     \\
     \\
    **Multi-round interaction full example**   \\
    \texttt{\`}\texttt{\`}\texttt{\`}
    \{example\}
    \texttt{\`}\texttt{\`}\texttt{\`} \\
    
     \\
    **Notes**   \\
    1. **You must generate the position markers** in accordance with the requirements stated above (<thought></thought>; <tool\_call></tool\_call> or <answer></answer>).  
    2. In particular, do not forget to generate the closing tags: </thought>, </tool\_call>, OR </answer> \\
    3. You must not generate extra angle bracket position markers. \\

    \bottomrule
    \end{tabularx}
    \caption{ReAct system prompt.}
    \label{table:prompt_react}
\end{table*}
\begin{table*}[ht]
\renewcommand{\arraystretch}{1.0}
\small
\centering
    \begin{tabularx}{\textwidth}{X}
    \toprule
    \textbf{Prompt for generating experience triplets.} \\
    \midrule

    \# Questions for Students to Solve \\
    \{question\} \\
     \\
    \# Standard Answer for the Question\\
    \{answer\} \\
     \\
    \# This is a complete trajectory that ultimately got the correct answer as your reference: \\
    \texttt{\`}\texttt{\`}\texttt{\`}\{true\_traj\}\texttt{\`}\texttt{\`}\texttt{\`} \\
     \\
    \# This is a complete trajectory that ultimately got the wrong answer: \\
    \texttt{\`}\texttt{\`}\texttt{\`}\{false\_traj\}\texttt{\`}\texttt{\`}\texttt{\`} \\
     \\
     \\
    \# Pattern of All Trajectories: \\
    Question, R1, O1, R2, O2, R3, O3, R4... \\
     \\
    \# Instructions and Your Task \\
    1. Define a STEP as R\_i+O\_i, but the last STEP only has R\_N \\
    2. Each R is a student's response, attempting to call tools to further solve the problem, but the second trajectory with wrong answer always has some issues \\
    3. Your core task is to answer this question for each STEP: \texttt{\`}\texttt{\`}\texttt{\`}In order to avoid the final error, if guidance is provided after this STEP ends, what should be done to make the agent perform better?\texttt{\`}\texttt{\`}\texttt{\`} \\
    4. Of course, a complete guidance is a triplet <student's current state, reason why this STEP leads to the final error, what to say before the next STEP to improve the current state> \\
        \texttt{\`}\texttt{\`}\texttt{\`}Explanation of the triplet: \\
        \quad- Student's current state: A relatively general description, introducing what the student saw and what they did. The description does not involve error attribution, focuses on objectively stating the student's behavior, and does not evaluate whether the behavior is good/bad or right/wrong \\
        \quad- Reason why this STEP leads to the final error: Unlike the current state, this part explicitly points out what mistake the student made in this STEP \\
        \quad- What to say before the next STEP to improve the current state: Based on the errors mentioned above, provide specific guidance that will help the student perform better in the next STEP if they follow it. Of course, do not directly tell the student the answer! \\
        \texttt{\`}\texttt{\`}\texttt{\`} \\
    5. The guidance opinion in the triplet generated for STEP\_i will be concatenated after O\_i, which means the student can see it before generating R\_i+1 \\
    6. Not every STEP necessarily needs guidance, you can skip after analysis, but since the trajectory is wrong, **there must be at least one STEP that has issues and can be summarized into a triplet** \\
    7. Finally, briefly summarize what three good pieces of advice could be given before working on this problem \\
    8. **!!Must Note!!** The total number of rounds you analyze in the trajectory is **\{step\_num\}**, you must generate the corresponding number of STEPs before you can continue to generate TOTAL! \\
     \\
    \# Output Format (strictly follow the markdown format I give you) \\
    \texttt{\`}\texttt{\`}\texttt{\`} \\
    \# STEP 1: \\
    \#\# Analysis \\
    - Write analysis content here \\
    \#\# Triplet \\
    (If there is no error, directly write "- None", do not generate a triplet when there is no error) \\
    - Student's current state: Write current state here \\
    - Reason why this STEP leads to the final error: Write reason here \\
    - What to say before the next STEP to improve the current state: Write guidance here \\
     \\
    \# STEP 2: \\
    ... \\
    \texttt{\`}\texttt{\`}\texttt{\`} \\

    \bottomrule
    \end{tabularx}
    \caption{Prompt for generating experience triplets.}
    \label{table:prompt_generate_triplet}
\end{table*}
\begin{table*}[ht]
\renewcommand{\arraystretch}{1.0}
\small
\centering
    \begin{tabularx}{\textwidth}{X}
    \toprule
    \textbf{Prompt for iteratively generating topics.} \\
    \midrule

    A teacher is analyzing each step taken by students when solving complex problems. I will give you several ``behavior + mistake'' items that the teacher has summarized for students at certain steps, defined as: \\
    \texttt{\`}\texttt{\`}\texttt{\`} \\
    behavior: A relatively general description introducing what the student saw and then did. The description does not involve error attribution, focuses on objectively stating the student's behavior, and does not evaluate whether the behavior is good/bad or right/wrong \\
    mistake: The student is satisfied with preliminary information (such as ``Deputy Party Secretary'') and fails to realize the need to cross-reference multiple search results to extract a complete and accurate answer, particularly overlooking that the specific name ``xx xx xx'' has already appeared in the first entry of the third set of search results. \\
    \texttt{\`}\texttt{\`}\texttt{\`} \\
     \\
    \# Overall Overview \\
    Your goal is to give **each behavior + mistake** a **scenario-narrative label**. A label's description should be concise enough to clearly express the characteristics of the behavior and be reusable. \\
    For each given new behavior + mistake, you can choose one of the following three actions: \\
    1. Reuse: Do not change any current labels, and select an existing label for the new behavior (recognizing the existing classification) \\
    2. Create: Do not change existing labels, create a new label for the new behavior (existing classification is incomplete) \\
    3. Modify: Modify certain current labels, and assign that label to the new behavior (existing classification is inaccurate) \\
     \\
    \# Detailed Requirements \\
    1. Each label must be concise and clear, but needs to have certain semantic information that allows people to understand the characteristics of the current behavior + mistake without explanation. It should be at least a dozen or dozens of words (e.g., in the pattern of xxx: xxx xxx xx). \\
    2. There cannot be too many labels; each label should have distinguishability in scenario content. \\
    3. One label can correspond to multiple behaviors, so you must ensure their textual content is consistent. \\
    4. Use the given id as the unique identifier for behaviors. When outputting, you need to output the ids and labels of all existing behaviors and new behaviors. \\
    5. Try to keep the number of different labels balanced. \\
     \\
    \# List of Behaviors Already Given Labels \\
    \{exp\_list\} \\
     \\
    \# List of New Behaviors \\
    \{new\_exp\_list\} \\
     \\
    \# Output Format: \\
    \texttt{\`}\texttt{\`}\texttt{\`} \\
    \{output\_format\} \\
    \texttt{\`}\texttt{\`}\texttt{\`} \\
    
    \bottomrule
    \end{tabularx}
    \caption{Prompt for iteratively generating topics.}
    \label{table:prompt_generate_topic}
\end{table*}
\begin{table*}[ht]
\renewcommand{\arraystretch}{1.0}
\small
\centering
    \begin{tabularx}{\textwidth}{X}
    \toprule
    \textbf{Prompt for experience model topic selection stage.} \\
    \midrule

    \# Overall Instructions \\
    You are a teacher who is very good at guiding students to answer difficult questions
    Now, the student has made some attempts on the given problem, but there may be some non-absolute but potential issues, or even errors, in the **current step**.
    I will give you the student's multiple rounds of attempts from the beginning to the current state, where assistant represents the student's response, user represents the environment's return value, and may also include your previous guidance.
    In addition, a topic list full of your guidance experience is provided for you to choose from. \\
     \\
    \# Student's Multiple Rounds of Interaction \\
    \texttt{\`}\texttt{\`}\texttt{\`}\{history\}\texttt{\`}\texttt{\`}\texttt{\`} \\
     \\
    \# Candidate Topic List \\
    \texttt{\`}\texttt{\`}\texttt{\`}\{topic\_list\}\texttt{\`}\texttt{\`}\texttt{\`} \\
     \\
    \# Your Task \\
    1. You need to combine the student's current state and select **3** topics from the several potential error topics I give you. After you carefully state the reasons for selection, just output the idx of the selected topics. \\
    2. The student may not have actually made a mistake, but your subsequent guidance can prevent problems before they occur. \\
     \\
    \# Output Format \\
    Strictly follow the markdown format below for output \\
    \texttt{\`}\texttt{\`}\texttt{\`} \\
    \# Analysis of the Current Step \\
    Write your analysis here \\
    \# Selected Topic idx (separated by spaces) \\
    idx1 idx2 idx3 \\
    \texttt{\`}\texttt{\`}\texttt{\`} \\
     \\
    Output: \\

    \bottomrule
    \end{tabularx}
    \caption{Prompt for experience model topic selection stage.}
    \label{table:prompt_stage_one}
\end{table*}
\begin{table*}[ht]
\renewcommand{\arraystretch}{1.0}
\small
\centering
    \begin{tabularx}{\textwidth}{X}
    \toprule
    \textbf{Prompt for experience model guidance generation stage.} \\
    \midrule

    \# Overall Instructions \\
    You are a teacher who is very good at guiding students to answer difficult questions.
    Now, the student has made some attempts on the given problem, but there may be some non-absolute but potential issues, or even errors, in the **current step**.
    I will give you the student's multiple rounds of attempts from the beginning to the current state, where assistant represents the student's response, user represents the environment's return value, and may also include your previous guidance.
    In addition, you are given your previous rich teaching experience, which consists of guidance you provided for different problem steps. Your goal is to provide the most appropriate guidance for the current step. \\
     \\
    \# Student's Multiple Rounds of Interaction \\
    \texttt{\`}\texttt{\`}\texttt{\`}
    \{history\}
    \texttt{\`}\texttt{\`}\texttt{\`} \\
     \\
    \# Previous Rich Guidance Experience \\
    These guidances are based on the step problems you discovered in students, so each content is a triplet, including the student's behavior, the error, and your previous guidance \\
    \texttt{\`}\texttt{\`}\texttt{\`}
    \{topic\_kb\}
    \texttt{\`}\texttt{\`}\texttt{\`} \\
     \\
    \# Your Task and Requirements \\
    1. The question must have an answer. If the student thinks there is insufficient evidence, it must be because they haven't found the evidence. After careful analysis, provide your guidance for the student's current step, with the goal of helping the student actually answer the question correctly. \\
    2. Since this step may not necessarily be wrong, please carefully choose your wording to prevent your guidance from introducing bias. \\
    3. Your analysis must include a brief review of the **problem** that the student needs to solve, emphasizing the content of the problem to the student to prevent answering off-topic. \\
    4. The guidance you provide will be given to the student together with the tool call results after this step ends. \\
        \quad- If the student generates a tool call in this step, the guidance will be given to the student together with the tool return value \\
        \quad- If the student generates an answer in this step, the guidance will be given directly to the student, and the student will choose whether to answer again or continue to generate tool calls based on your guidance \\
    5. It is forbidden to find answers on behalf of the student, and it is forbidden to hint at what the answer is under any circumstances. **You are a teacher, not someone helping students cheat**. \\
    6. It is forbidden to provide **direct clues** to students. Your purpose is only to **guide**. \\
    7. Guidance should be clear and easy to understand. If necessary, you can encourage students to continue calling tools or switch tools. \\
    8. The guidance you provide should try to imitate the previous guidance patterns, don't improvise freely. \\
     \\
    \# Output Format \\
    Strictly follow the markdown format below for output \\
    \texttt{\`}\texttt{\`}\texttt{\`} \\
    \# Analysis combining student's current behavior and previous experience to provide appropriate guidance for the present moment \\
    Write your detailed analysis here \\
    \# Guidance Content \\
    Write your guidance content here \\
    \texttt{\`}\texttt{\`}\texttt{\`} \\
     \\
    Output:\\

    \bottomrule
    \end{tabularx}
    \caption{Prompt for experience model guidance generation stage.}
    \label{table:prompt_stage_two}
\end{table*}
\begin{table*}[ht]
\renewcommand{\arraystretch}{1.0}
\small
\centering
    \begin{tabularx}{\textwidth}{X}
    \toprule
    \textbf{Prompt for generating guidance experience directly without referencing the experience base.} \\
    \midrule

    \# Overall Instructions \\
    You are a teacher who is very good at guiding students to answer difficult questions.
    Now, the student has made some attempts on the given problem, but there may be some non-absolute but potential issues, or even errors, in the **current step**.
    I will give you the student's multiple rounds of attempts from the beginning to the current state, where assistant represents the student's response, user represents the environment's return value, and may also include your previous guidance.
    Your goal is to provide the most appropriate guidance for the current step. \\
     \\
    \# Student's Multiple Rounds of Interaction \\
    \texttt{\`}\texttt{\`}\texttt{\`}
    \{history\}
    \texttt{\`}\texttt{\`}\texttt{\`} \\
     \\
    \# Your Task and Requirements \\
    1. The question must have an answer. If the student thinks there is insufficient evidence, it must be because they haven't found the evidence. After careful analysis, provide your guidance for the student's current step, with the goal of helping the student actually answer the question correctly. \\
    2. Since this step may not necessarily be wrong, please carefully choose your wording to prevent your guidance from introducing bias. \\
    3. Your analysis must include a brief review of the **problem** that the student needs to solve, emphasizing the content of the problem to the student to prevent answering off-topic. \\
    4. The guidance you provide will be given to the student together with the tool call results after this step ends \\
        \quad- If the student generates a tool call in this step, the guidance will be given to the student together with the tool return value \\
        \quad- If the student generates an answer in this step, the guidance will be given directly to the student, and the student will choose whether to answer again or continue to generate tool calls based on your guidance \\
    5. It is forbidden to find answers on behalf of the student, and it is forbidden to hint at what the answer is under any circumstances. **You are a teacher, not someone helping students cheat**. \\
    6. It is forbidden to provide **direct clues** to students. Your purpose is only to **guide**. \\
    7. Guidance should be clear and easy to understand. If necessary, you can encourage students to continue calling tools or switch tools. \\
     \\
    \# Output Format \\
    Strictly follow the markdown format below for output \\
    \texttt{\`}\texttt{\`}\texttt{\`} \\
    \# Analysis combining student's current behavior to provide appropriate guidance for the present moment \\
    Write your detailed analysis here \\
    \# Guidance Content \\
    Write your guidance content here \\
    \texttt{\`}\texttt{\`}\texttt{\`} \\
     \\
    Output: \\
    
    \bottomrule
    \end{tabularx}
    \caption{Prompt for generating guidance experience directly without referencing the experience base.}
    \label{table:prompt_direct_guide}
\end{table*}
\begin{table*}[ht]
\renewcommand{\arraystretch}{1.0}
\small
\centering
    \begin{tabularx}{\textwidth}{X}
    \toprule
    \textbf{Prompt for model-based guidance triggering decision.} \\
    \midrule

    \# Overall Instructions \\
    You are a teacher who is very good at guiding students to answer difficult questions.
    Now, the student has made some attempts on the given question, but there may be some non-absolute yet potential issues or even errors in the current step.
    I will provide you with the student's multi-turn attempts from the beginning to the current point, where ``assistant'' represents the student's responses and "user" represents environmental feedback, which may also include the teacher's prior guidance.
    Your purpose is to determine whether guidance is needed at the current step. \\
     \\
    \# Student's Multi-turn Interactions \\
    \texttt{\`}\texttt{\`}\texttt{\`}
    \{history\}
    \texttt{\`}\texttt{\`}\texttt{\`} \\
     \\
    \# Your Task and Requirements \\
    Analyze whether the student's behavior in the current step has potential issues, and provide a signal indicating whether guidance is needed. \\
     \\
    \# Output Format \\
    Strictly follow the markdown format below for output \\
    \texttt{\`}\texttt{\`}\texttt{\`} \\
    \# Your Analysis \\
    Your analysis content here \\
     \\
    \# Whether to Provide Guidance \\
    yes/no \\
    \texttt{\`}\texttt{\`}\texttt{\`} \\
     \\
    Output: \\
    
    \bottomrule
    \end{tabularx}
    \caption{Prompt for model-based guidance triggering decision.}
    \label{table:prompt_model_as_trigger}
\end{table*}
\begin{table*}[ht]
\renewcommand{\arraystretch}{1.0}
\small
\centering
    \begin{tabularx}{\textwidth}{X}
    \toprule
    \textbf{Prompt for retrieving and directly using the experience base.} \\
    \midrule

    Your current situation resembles a mistake made by a previous student. That student's behavior was: \\
    \{behavior\} \\
     \\
    Their mistake was: \\
    \{mistake\} \\
     \\
    My guidance to them was: \\
    \{guidance\} \\
     \\
    Learn from this error, examine your own behavior, and proceed with better decision-making. \\
    
    \bottomrule
    \end{tabularx}
    \caption{Prompt for retrieving and directly using the experience base.}
    \label{table:prompt_emb}
\end{table*}

\begin{table*}[ht]
\renewcommand{\arraystretch}{1.0}
\small
\centering
    \begin{tabularx}{\textwidth}{X}
    \toprule
    \textbf{Prompt for evaluation.} \\
    \midrule

    You are an evaluation assistant. Please determine if the predicted answer is equivalent to the labeled answer.
    As long as the Labeled Answer is contained within the Predicted Answer, it is considered correct, even if additional, more detailed explanations are included.\\
    You should focus on whether the Predicted Answer truly answers the question correctly.\\
    \\    
    \# Question: \\    
    \{question\}\\    
    \\    
    \# Labeled Answer: \\    
    \{correct\_answer\}\\    
    \\    
    \# Predicted Answer: \\    
    \{response\}\\    
    \\    
    Respond **only with** "Correct" or "Incorrect", no other tokens.\\

    \bottomrule
    \end{tabularx}
    \caption{Prompt for evaluation.}
    \label{table:prompt_eval}
\end{table*}

\end{document}